%% file: main.tex
\documentclass[nohyperref]{article}

\usepackage{microtype}
\usepackage{graphicx}
\usepackage{subfigure}
\usepackage{booktabs} 
\usepackage{subfiles}
\usepackage{float}

\newcommand{\ve}[1]{\mathbf{#1}}
\newcommand{\ves}[1]{\boldsymbol{#1}}

\newcommand{\dd}{\text{d}}
\newcommand{\ddt}{\frac{\text{d}}{\text{d}t}}
\newcommand{\myequation}{\begin{equation}}
\newcommand{\myendequation}{\end{equation}}
\newcommand{\appropto}{\mathrel{\vcenter{
  \offinterlineskip\halign{\hfil$##$\cr
    \propto\cr\noalign{\kern2pt}\sim\cr\noalign{\kern-2pt}}}}}
\let\[\myequation
\let\]\myendequation

\newcommand{\J}{J}

\newcommand{\vfb}{\ve{v}^{\mathrm{fb}}}
\newcommand{\vff}{\ve{v}^{\mathrm{ff}}}

\newcommand{\tauv}{\tau_v}
\newcommand{\taue}{\tau_{\epsilon}}
\newcommand{\sss}{\mathrm{ss}}

\newcommand{\veps}{\ves{\epsilon}}
\newcommand{\vex}{\ves{\xi}}
\newcommand{\vrl}{\ve{r}_L}
\newcommand{\vv}{\ve{v}}

\newcommand{\dvvt}{\Delta \tilde{\vv}}

\newcommand{\vvbsi}{\bar{\vv}_{\sss,i}}
\newcommand{\vvt}{\tilde{\vv}}
\newcommand{\vvbs}{\bar{\vv}_{\sss}}
\newcommand{\vr}{\ve{r}}
\newcommand{\vu}{\ve{u}}
\newcommand{\vus}{\ve{u}_{\mathrm{ss}}}

\newcommand{\vut}{\tilde{\vu}}
\newcommand{\vub}{\bar{\vu}}
\newcommand{\vubs}{\bar{\vu}_{\sss}}
\newcommand{\vuhp}{\vu_{\mathrm{hp}}}
\newcommand{\vulp}{\vu_{\mathrm{lp}}}
\newcommand{\vt}{\ves{\theta}}
\newcommand{\uint}{\ve{u}^{\mathrm{int}}}
\newcommand{\calH}{\mathcal{H}}
\newcommand{\calL}{\mathcal{L}}
\newcommand{\Js}{J_{\mathrm{ss}}}

\newcommand{\talpha}{\tilde{\alpha}}
\newcommand{\Rs}{R_{\mathrm{ss}}}
\newcommand{\vgamma}{\ves{\gamma}}
\newcommand{\vrltrue}{\vrl^{\mathrm{true}}}

\newcommand{\snorm}[1]{\lVert#1\rVert}
\newcommand{\pder}[2]{\frac{\partial#2}{\partial #1}}

\newcommand{\der}[2]{\frac{\mathrm{d}#2}{\mathrm{d} #1}}

\newcommand{\bbE}{\mathbb{E}}

\usepackage{hyperref}

\usepackage[accepted]{icml2022}

\usepackage{amsmath}
\usepackage{amssymb}
\usepackage{mathtools}
\usepackage{amsthm}

\usepackage[capitalize,noabbrev]{cleveref}
\usepackage{comment}

\theoremstyle{plain}
\newtheorem{theorem}{Theorem}
\newtheorem{proposition}[theorem]{Proposition}
\newtheorem{lemma}[theorem]{Lemma}

\theoremstyle{definition}

\newtheorem{condition}{Condition}
\theoremstyle{remark}

\usepackage[textsize=tiny]{todonotes}

\icmltitlerunning{Minimizing Control for Credit Assignment with Strong Feedback}

\begin{document}

\twocolumn[
\icmltitle{Minimizing Control for Credit Assignment with Strong Feedback}



\icmlsetsymbol{equal}{*}

\begin{icmlauthorlist}
\icmlauthor{Alexander Meulemans}{equal,ini}
\icmlauthor{Matilde Tristany Farinha}{equal,ini}
\icmlauthor{Maria Cervera}{equal,ini} \\
\icmlauthor{João Sacramento}{ini}
\icmlauthor{Benjamin F. Grewe}{ini}
\end{icmlauthorlist}

\icmlaffiliation{ini}{Institute of Neuroinformatics, University of Zürich and ETH Zürich, Switzerland}

\icmlcorrespondingauthor{Alexander Meulemans}{ameulema@ethz.ch}

\icmlkeywords{Machine Learning, ICML}

\vskip 0.3in
]



\printAffiliationsAndNotice{\icmlEqualContribution} 

\begin{abstract}
    The success of deep learning ignited interest in whether the brain learns hierarchical representations using gradient-based learning.  
    However, current biologically plausible methods for gradient-based credit assignment in deep neural networks need infinitesimally small feedback signals, which is problematic in biologically realistic noisy environments and at odds with experimental evidence in neuroscience showing that top-down feedback can significantly influence neural activity.
    Building upon deep feedback control (DFC), a recently proposed credit assignment method, we combine \textit{strong} feedback influences on neural activity with gradient-based learning and show that this naturally leads to a novel view on neural network optimization.
    Instead of gradually changing the network weights towards configurations with low output loss, weight updates gradually minimize the amount of feedback required from a controller that drives the network to the supervised output label.
    Moreover, we show that the use of strong feedback in DFC allows learning forward and feedback connections simultaneously, using learning rules fully local in space and time.
    We complement our theoretical results with experiments on standard computer-vision benchmarks, showing competitive performance to backpropagation as well as robustness to noise.
    Overall, our work presents a fundamentally novel view of learning as control minimization, while sidestepping biologically unrealistic assumptions.
\end{abstract}

\section{Introduction}
\label{intro}

The error backpropagation (BP) method \citep{rumelhart1986learning,werbos1982applications} has emerged as the method of choice for training deep artificial neural networks, due to its efficient computation of gradients, which is crucial for learning in high-dimensional parameter spaces \citep{lecun2015deep}. Although biological networks in the neocortex form similar hierarchies \citep{yamins2016using}, it is not clear yet if they follow similar principles to implement credit assignment (CA), i.e., determining how to change synaptic connectivity strengths to get closer to some desired output. Several key properties of the BP method are highly incompatible with cortical networks \citep{crick1989recent, lillicrap2020backpropagation}. For example, synaptic plasticity in biological neurons is local in space and time and tightly coupled to neural activity. By contrast, artificial neural networks trained with BP process activity and weight update signals separately and in distinct phases while requiring exactly the same synaptic weights in the forward and feedback pathways.

To couple synaptic plasticity directly to neural activity, recent work designed cortical network models that propagate CA signals by leveraging the network dynamics \citep{scellier2017equilibrium, whittington2017approximation, sacramento2018dendritic, meulemans2021credit} or using multiplexed neural signals \citep{payeur2020burst}. However, these methods require tightly coordinated plasticity mechanisms, possibly through the use of distinct training phases, and it remains unclear whether cortical networks can exhibit these levels of coordination. Moreover, to approximate gradient-based optimization these methods consider the \textit{weak-feedback} limit where, to avoid interference, feedback cannot significantly alter the feedforward processing of the network. This weak-feedback assumption is problematic in biologically realistic noisy environments, and is in stark contrast with biological observations showing that top-down feedback can significantly affect forward processing  \citep{jordan2020opposing, keller2020feedback}. 

To date, several studies used strong feedback control signals in combination with error-based learning rules to drive the network activity to track a reference output \citep{gilra2017predicting, alemi2017learning, deneve2017brain, bourdoukan2015enforcing}. Although these models use strong feedback influences on neural activity, they have only been successfully applied to train single-layer recurrent neural networks with fixed feedback weights and it remains unclear whether they can be extended for training deep neural networks. Furthermore, even though these methods are capable of reducing the output loss during training, it remains an open question whether they extract gradient information to enable efficient learning. Hence, gradient-based CA theories for training deep neural networks with strong feedback influences and without the need for tightly coordinated plasticity mechanisms are still lacking. 

Building upon Deep Feedback Control (DFC) \citep{meulemans2021credit}, we introduce Strong-DFC, a gradient-based approach to CA in deep neural networks that makes use of strong feedback influences on neural activity. Like DFC, we use a feedback controller to drive the network to a desired output target and update the network weights with a learning rule fully local in space and time which leverages the dynamic change in neural activity. Unlike DFC, we allow for strong feedback influences by taking the supervised label as the desired output target instead of weakly nudging the current network output towards the supervised label. This new setting results in significant feedback influence on the neural activities, as the initial network output is far from the supervised label at the beginning of training. 

Interestingly, because the output targets remain invariant throughout training, Strong-DFC naturally leads to a fundamentally different perspective on neural network optimization. As the controller drives the network to the supervised output labels at each point in training, the goal of optimizing the network parameters shifts from minimizing the supervised loss towards minimizing the amount of feedback control needed to reach the supervised output labels. To formalize this \textit{minimizing control} perspective we introduce a surrogate loss that quantifies the amount of control feedback, and show that, under flexible conditions on the feedback connectivity, the Strong-DFC updates follow its negative gradient. Intriguingly, Strong-DFC enables us to learn the forward and feedback weights simultaneously, overcoming the two-phase requirement of DFC, while inheriting the close connection to recently proposed dendritic compartment models of cortical pyramidal neurons equipped with voltage-dependent plasticity rules \citep{sacramento2018dendritic}. Finally, we complement our theoretical findings with experimental results on standard computer vision benchmarks, showing that Strong-DFC provides principled CA in deep neural networks using strong feedback influences without the need for tightly coordinated plasticity mechanisms.

\section{Background}
\label{background}
We revisit DFC \citep{meulemans2021credit} to investigate CA with strong feedback influence on neural activities. The control framework used by DFC was designed for feedforward networks with continuous dynamics. More specifically, DFC considers a multilayer network with the following dynamics (for $1\leq i \leq L)$:
\vspace{-0.05cm}
\begin{align}\label{eq:network_dynamics}
    \tau_v \dot{\ve{v}}_i(t) &= -\ve{v}_i(t) + W_i\phi\big(\ve{v}_{i-1}(t)\big) + Q_i\ve{u}(t),
\end{align}

where $\ve{v}_i$ is a vector of pre-nonlinearity activations in layer $i$, $\dot{\ve{v}}_i$ its time derivative, $W_i$ the forward weights, $\phi$ a smooth nonlinear activation function, $Q_i$ the feedback weights, and $\tau_v$ the neural time constant. We define $\vr_i = \phi(\vv_i)$ as the post-nonlinearity activity of layer $i$. Note that without feedback influences ($\vu=0$), the steady state of the above dynamics corresponds to a feedforward neural network, as the input $\vr_0$ remains constant over time.

The feedback signal $\ve{u}$ is computed by a proportional integral feedback controller, using the output error $\ve{e}(t) = \vr_L^* - \ve{r}_L(t)$, and serves to push the network to a desired output target $\ve{r}_L^*$:
\begin{align} \label{eq:controller_dynamics}
    &\ve{u}(t) = \ve{u}^{\text{int}}(t) + k \ve{e}(t), &
    \tau_u \dot{\ve{u}}^{\text{int}}(t) = \ve{e}(t) - \alpha \ve{u}^{\text{int}}(t),
\end{align}
where $k$ is the proportional control constant and $\alpha$ the leakage constant limiting the magnitude of $\uint$. DFC weakly nudges the network output in the direction of lower loss, since the output target $\vrl^*$ is defined as:
\begin{align} \label{eq:nudged_output_target}
    \vrl^* = \vrl^- - \lambda \left. \pder{\vrl}{\calL}\right\vert_{\vrl=\vrl^-}^T,
\end{align}
with $\vrl^-$ the steady state of the output without the presence of a controller, $\calL$ the loss function, and $\lambda$ the nudging strength. The theoretical results linking DFC to a variant of Gauss-Newton optimization consider the \textit{weak-feedback} regime of $\lambda \rightarrow 0$ \citep{meulemans2021credit}. Note that we cannot enforce \textit{strong feedback} in this theoretical framework by keeping the small nudging factor $\lambda \rightarrow 0$ and making $Q_i$ arbitrarily large, as the limit considered by \citet{meulemans2021credit} relies on vanishing feedback influence on neural activity, similar to other seminal work in dynamic CA methods that approximate gradient-based optimization \citep{sacramento2018dendritic, scellier2017equilibrium, whittington2017approximation}. DFC then updates the feedforward weights with the following plasticity rule that is fully local in space and time:
\begin{align}\label{eq:W_dynamics}
    \tau_W  \dot{W}_i (t) = \big(\phi(\ve{v}_i(t)) - \phi(W_i\ve{r}_{i-1}(t))\big)\ve{r}_{i-1}(t)^T.
\end{align}

\begin{figure*}[t!]
\centering
\includegraphics[width=0.8\textwidth]{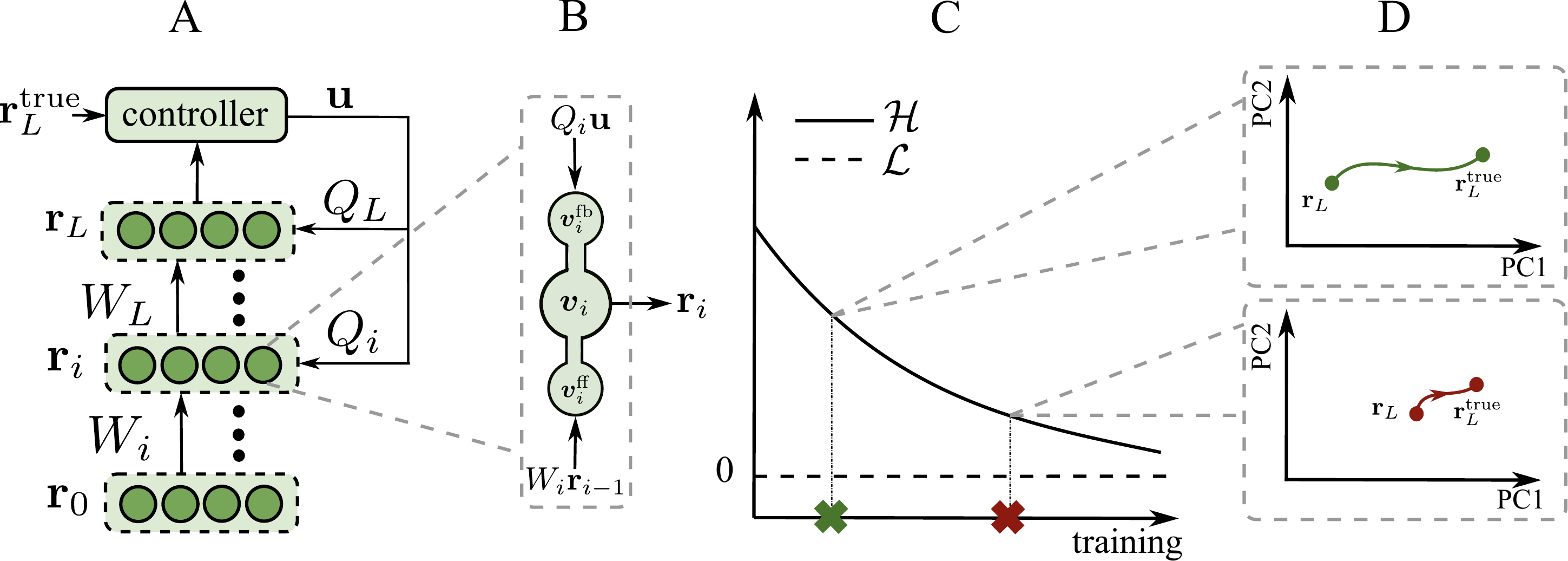}
\vspace{-0.3cm}
\caption{
\textbf{Strong-DFC trains deep neural networks by minimizing control.} (A) Schematic of a feedforward network trained with Strong-DFC, where feedback weights $Q$ are uncoupled from forward weights $W$, and used to carry a control signal $\mathbf{u}$ from the controller to all hidden neurons. (B) Neurons are multi-compartmental and allow integrating separate forward drive $v_i^{\text{ff}}$ and feedback control $v_i^{\text{fb}}$ signals. (C) Since the supervised target $\mathbf{r}_L^{\text{true}}$ is always reached whenever the controller is on, the training loss $\mathcal{L}$ is equal to zero throughout, and training can be better understood as the minimization of a surrogate loss $\mathcal{H}$ describing the amount of control required. (D) Early in training (top) the controller feedback strongly influences the neural activity to drive the network output to the desired target, and training minimizes the amount of control required (bottom). Figure A-B adapted from \citet{meulemans2021credit}.}
\label{fig:neuron_and_dynamics_double_column}
\end{figure*}

Using a multi-compartment model of a neuron (Fig. \ref{fig:neuron_and_dynamics_double_column}B), one can interpret this learning rule as a delta rule that uses the difference between the outgoing firing rate of the neuron $\phi(\vv_i)$ and the estimated firing rate based only on the feedforward input of the neuron $\phi(\vff_i)=\phi(W_i \vv_{i-1})$ \citep{urbanczik2014learning}. Standard DFC learns its feedback weights $Q_i$ in a separate `sleep' phase with a simple anti-Hebbian learning rule, making use of the feedback compartment of the multi-compartment neuron, $\vfb_i=Q_i\vu$. 
Note that DFC assumes a separation of timescales between the plasticity dynamics \eqref{eq:W_dynamics} and the network dynamics (\ref{eq:network_dynamics}-\ref{eq:controller_dynamics}), such that the weights $W_i$ and $Q_i$ can be considered constant on the timescale of the network-controller interactions.

\section{Training neural networks with strong feedback by minimizing control} \label{sec:minimizing_control}

Here, we introduce our \textit{minimizing control} framework that allows for gradient-based CA with strong feedback influences in deep neural networks. We use the feedback controller \eqref{eq:controller_dynamics} to strongly drive the network towards the supervised output label at each point during training. Consequently, the focus of optimizing the network parameters shifts from minimizing the training loss $\calL$ towards minimizing the amount of feedback control needed to reach the supervised output label. We quantify the magnitude of the control input with a surrogate loss $\calH$, and show that under flexible constraints on the feedback connectivity, the Strong-DFC updates perform gradient descent on this new loss function. Finally, we show that by minimizing $\calH$, we implicitly minimize $\calL$ as well. Throughout this section, we assume the network is stable, on which we elaborate at the end of the section. Although the theoretical results shown here are tailored towards the proportional integral controller \eqref{eq:controller_dynamics}, these can easily be extended to other controller types such as a pure proportional controller (App. \ref{app:other_controllers}).

\vspace{-0.3cm}
\paragraph{Strong-DFC.} We adapt DFC to allow for strong feedback influence on neural activity by setting the output target $\vrl^*$ to the supervised output label $\vrl^{\mathrm{true}}$, instead of weakly nudging it towards values of lower loss \eqref{eq:nudged_output_target} (see Fig. \ref{fig:neuron_and_dynamics_double_column}). This new setting results in strong feedback influence, as the network output without feedback is far from the supervised labels at the beginning of training. 
Furthermore, as the existing theory on DFC relies on the assumption that the feedback only infinitesimally changes the network activity, it is no longer valid in this new setting of DFC with strong feedback (Strong-DFC). 

\paragraph{Learning as control minimization.} Here, we propose a novel view on neural network optimization that justifies strong feedback influences on neural activity. In Strong-DFC, at every stage of training the feedback controller drives the network to the supervised label $\vrltrue$, resulting in zero training loss at steady state. Hence, instead of gradually changing the network weights towards configurations with low output loss, the goal of optimizing the network parameters is now to gradually reduce the amount of feedback required from the controller to reach the supervised output label (see Fig. \ref{fig:neuron_and_dynamics_double_column}C-D). At the individual neuron level, one can interpret this new perspective on optimization as each neuron minimizing its feedback input by altering its feedforward weights.

We formalize this notion of \textit{minimizing control} by introducing a surrogate loss function representing the magnitude of the feedback control input: 
\begin{align}
\label{eq:surrogate-loss}
    \mathcal{H} = \frac{1}{2}\sum_b\|Q\mathbf{u}^{(b)}_{\mathrm{ss}}\|_2^2,
\end{align}
with $Q\triangleq[Q^T_1,...,Q^T_L]^T$ the concatenated feedback weights and $\mathbf{u}^{(b)}_{\mathrm{ss}}$ the steady-state control feedback of a data point $b$. The control feedback $\ve{u}_{\sss}$ depends on the feedforward weights $W_i$ through the network and controller dynamics (Eq. \ref{eq:network_dynamics}-\ref{eq:controller_dynamics}). Using the implicit function theorem, in Theorem \ref{theorem:grad_H} we compute the total derivative of $\calH$ w.r.t. $W_i$ and show that, under flexible constraints on the feedback weights, it results in a local update in space and time that aligns with a simplified version of the Strong-DFC weight updates (full proof in App. \ref{app:proof1}).

\begin{theorem}\label{theorem:grad_H}
Assuming $\Js Q + \talpha I$ is invertible, the total derivative of $\calH$ w.r.t. $W_i$ is given by:
\begin{align}\label{eq:total_grad_H}
    \der{W_i^T}{\calH} = -\sum_b J_{i, \sss}^{(b)T} \left(\Js^{(b)} Q + \talpha I\right)^{-T}Q^TQ  \vus^{(b)} \vr_{i-1, \sss}^{(b)T},
\end{align}
with $J_{i, \sss} \triangleq \pder{\vv_i}{\vrl}$ and $J_{\mathrm{ss}} \triangleq [\frac{\partial \vr_L}{\partial \vv_1}, ...,\frac{\partial \vr_L}{\partial \vv_L}]$, both evaluated at steady state, and $\talpha \triangleq \frac{\alpha}{1 + k \alpha}$.

Furthermore, assuming Condition \ref{con:col_space} holds, the network is stable, $\alpha \rightarrow 0$, and $\J_{\mathrm{ss}}Q + \talpha I$ is invertible, then the following steady-state (ss) updates for the forward weights, averaged over the batch samples,
\begin{align}\label{eq:update_W_ss}
    \Delta W_{i, \mathrm{ss}} = \eta (\ve{v}_{i,\mathrm{ss}} - \ve{v}_{i,\mathrm{ss}}^{\mathrm{ff}})\ve{r}_{i-1,\mathrm{ss}}^T\,\,,
\end{align}

\vspace{-0.3cm}
with $\eta$ the stepsize, align with the negative gradient $-\frac{\dd \mathcal{H}}{\dd W_i}$.
\end{theorem}

\begin{condition}\label{con:col_space}
The column space of the concatenated feedback weights $Q$ is equal to the row space of $J_{\mathrm{ss}} \triangleq [\frac{\partial \vr_L}{\partial \vv_1}, ...,\frac{\partial \vr_L}{\partial \vv_L}]\big \vert_{\vv_i = \vv_{i,\mathrm{ss}}}$, the network Jacobian at steady state.
\end{condition}

Theorem \ref{theorem:grad_H} links Strong-DFC to gradient descent on a surrogate loss function $\calH$. Although the steady-state updates \eqref{eq:update_W_ss} of Theorem \ref{theorem:grad_H} are not exactly equal to the DFC updates \eqref{eq:W_dynamics}, previous work showed that it is a good approximation (see App.~\ref{app:nonlinearity_update}-\ref{app:steady_state_updates} and \citealp{meulemans2021credit}).  

\paragraph{Flexible feedback connectivity constraint.}
Condition~\ref{con:col_space} is a flexible connectivity constraint on the feedback weights that represents the requirement that the feedback controller influences the network activity in the most efficient manner (see App. \ref{app:col_condition}). More specifically, if Condition \ref{con:col_space} is not satisfied, the feedback input $Q\vus$ is partially lost in the nullspace of the network and hence the same influence on the network output could be reached with a smaller feedback input. Interestingly, Condition \ref{con:col_space} generalizes the feedback requirements of commonly used learning rules for approximating gradient-based optimization methods. For example, backpropagation and its approximations require $Q=J^T$ \citep{rumelhart1986learning, lansdell2019learning, akrout2019deep} and common variants of target propagation require the pseudoinverse $Q=J^{\dagger}$ \citep{meulemans2020theoretical}, both satisfying Condition \ref{con:col_space}. 
More generally, Condition \ref{con:col_space} implies that the feedback weights of Strong-DFC do not necessarily need to align with the forward pathway or its inverse to provide gradient-based CA by following the negative gradient of $\calH$. 
Remarkably, the fundamentally distinct theoretical framework of standard DFC gives rise to a similar condition on $Q$, however with $J$ evaluated at the feedforward activities without feedback present instead of the steady-state activities \citep{meulemans2021credit}.

\paragraph{Minimizing $\calH$ leads to minimizing $\calL$.}
Theorem \ref{theorem:grad_H} links Strong-DFC to gradient descent on the surrogate loss function $\mathcal{H}$. Hence, to show that Strong-DFC minimizes the original loss function $\calL$, we need to connect $\calH$ to $\calL$. Proposition \ref{proposition:min_H_min_L} shows that if $\calH$ is minimized to zero, $\calL$ is minimized to zero as well, which is trivial to prove as $\calH=0$ translates to all training samples reaching exactly their supervised output label without help from the controller. 

\begin{proposition}
\label{proposition:min_H_min_L}
For all $\mathcal{L}$ that obey $\mathcal{L}(\ve{x}, \ve{x})=0$, we have that
\begin{align*}
    \mathcal{H} = \sum_b\|Q\mathbf{u}^{(b)}_{\mathrm{ss}}\|_2^2 = 0 \iff \sum_b\mathcal{L}(\mathbf{r}_L^{-(b)}, \mathbf{r}_L^{\mathrm{true,}(b)}) = 0,
\end{align*}
with $\vrl^-$ the steady state of the output without the presence of a controller.
\end{proposition}

Recent work in the field of deep learning has shown that, for overparameterized networks, it is possible to minimize the training loss $\calL$ to zero while reaching good generalization \citep{belkin2019reconciling}. We show empirically in Section \ref{experiments} that Strong-DFC can minimize $\calH$, and hence $\calL$ to zero, thereby validating the minimizing control framework for optimizing neural networks.

\paragraph{Stability of Strong-DFC.} So far, we assumed that the network and controller dynamics reach their steady state. However, this is not a given, as the feedback interaction could lead to unstable loops causing the network to diverge. Adapting the results of \citet{meulemans2021credit}, we investigate the local stability of a simplified version of Strong-DFC that uses integral control and assumes a separation of timescales $\tauv \ll \tau_u$. At steady state, the Jacobian matrix of this dynamical system is equal to $-(\Js Q + \alpha I)$, resulting in the following stability condition on the feedback weights (see App. \ref{app:stability}).

\begin{condition}\label{con:stability}
Given the network Jacobian evaluated at steady state $\Js$, the real parts of the eigenvalues of $-\Js Q$ are all below $\alpha$.
\end{condition}

In App. \ref{app:stability}, we compute the Jacobian matrix of the full dynamical system representing Strong-DFC with proportional integral control and with no assumptions on a separation of timescales. However, a local stability analysis on this system does not reveal clearly interpretable stability conditions for the feedback weights. Therefore, we monitor instead the stability of the full system empirically.

\section{A single-phase learning scheme for forward and feedback weights} \label{sec:fb_learning}
Conditions \ref{con:col_space} and \ref{con:stability} highlight the importance of the feedback weights for learning and stability, respectively. As training causes the forward weights and thus the network Jacobian $\Js$ to change, the set of feedback connectivity patterns satisfying Conditions \ref{con:col_space} and \ref{con:stability} shift during learning as well. To keep the feedback weights compatible with the changing network, we learn the feedback weights together with the forward weights.

Here, we introduce plasticity dynamics for the feedback weights that make it possible to update both forward and feedback weights simultaneously in a single phase, in contrast to standard DFC and other methods that need two separate phases. Our new feedback plasticity dynamics leverage two essential properties of Strong-DFC. (i) The strong feedback influences make it possible to have noise in the dynamics while learning the forward weights, while keeping a good signal-to-noise ratio of the learning signals. (ii) The conditions on the feedback weights for learning and stability required for our minimizing control framework consider the network activity at steady state. Hence, the feedback weights should be learned at the same steady state used for training the forward weights. Leveraging the noise intrinsically available to neurons, we use a simple anti-Hebbian learning rule for the feedback weights to drive them towards satisfying Conditions \ref{con:col_space} and \ref{con:stability}.

\paragraph{Noisy neural dynamics.} Our feedback weight plasticity dynamics make use of noise injected in the neural dynamics, which we model as follows:
\begin{align} \label{eq:noisy_dynamics_v}
    \tau_v \dot{\vv}_i(t) &= -\vv_i(t) + W_i \phi\big(\vv_{i-1}(t)\big) + Q_i \vu(t) + \sigma \veps_i(t),
\end{align}
where the noise $\veps$ represents exponentially filtered white noise with time constant $\taue$, i.e., an Ornstein-Uhlenbeck process, and $\sigma$ the standard deviation of the noise. Note that such Ornstein-Uhlenbeck processes can, for example, be obtained in the limit of many independent Poisson inputs with infinitesimal synaptic weights \citep{gerstner2014neuronal}.

\paragraph{An anti-Hebbian learning rule for $Q$.} 
The noise fluctuations that propagate through the network carry information about the network Jacobian $J$ and we can extract this information by correlating the noise fluctuations at the output with the noise fluctuations at each layer. Inspired by the feedback weight learning of standard DFC \citep{meulemans2021credit}, we instantiate this correlation technique with the following simple anti-Hebbian plasticity rule: 
\begin{align}\label{eq:Q_dynamics_scaled}
    \tau_Q \dot{Q}_i(t) &= -\left(1+\frac{\tau_v}{\taue}\right)^{L-i}\vfb_i(t)\vu_{\mathrm{hp}}(t)^T - \beta Q_i.
\end{align}
The feedback compartment $\vfb$ contains a part of the injected noise, $\vfb_i = Q_i\vu + \sigma \veps_i^{\mathrm{fb}}$, and the pre-synaptic plasticity signal $\vuhp$ is a high-pass filtered version of the incoming control signal $\mathbf{u}$, which extracts the output noise fluctuations.\footnote{We high-pass filter $\vu(t)$ by subtracting the exponentially averaged signal from it: $\vuhp(t) \triangleq \vu(t) - \vulp(t)$, with $\vulp \triangleq  \frac{1}{\tau_f}\int_0^t \exp\big(-\frac{1}{\tau_f}(t-t')\big)\vu(t')\dd t'$ and $\tau_f$ the filtering time constant.}
Our feedback weight plasticity dynamics \eqref{eq:Q_dynamics_scaled} introduce two important novelties with respect to DFC. (i) The high-pass filtering of $\vu$ makes it possible to learn the feedback weights using the noise fluctuations of $\vu$, while simultaneously letting the controller push the network towards the output target for learning the forward weights. (ii) The scaling factor $(1+\tauv/\taue)$ enables the network to learn an optimal feedback connectivity pattern, while having a finite neural time constant $\tau_v$, which introduces delays in the network. As the correlation signal decreases for increasing time delays between the pre- and postsynaptic noise fluctuations, this needs to be compensated by an appropriate scaling factor.

\paragraph{Feedback weights satisfy Conditions \ref{con:col_space} and \ref{con:stability}.}
Theorem \ref{theorem:fb_learning} shows that under simplifying conditions, the plasticity rule Eq. \eqref{eq:Q_dynamics_scaled} drives the feedback connectivity $Q$ to satisfy Conditions \ref{con:col_space} and \ref{con:stability} (full proof in App. \ref{app:fb_learning}).

\begin{theorem}\label{theorem:fb_learning}
Assume a separation of timescales $\tauv, \taue \ll \ \tau_u \ll \tau_f \ll \tau_Q$, integral control ($k_p=0$), stable network dynamics, and small noise perturbations $\sigma \ll \snorm{\vv_i}$. Then, for a fixed data sample, the feedback plasticity dynamics \eqref{eq:Q_dynamics_scaled} let the first moment of $Q$ converge approximately towards
\begin{align}\label{eq:fb_alignment}
    \mathbb{E}[Q_{\sss}] \approx \Js^T M,
\end{align}
with $M$ a positive definite symmetric matrix, which satisfies Conditions \ref{con:col_space} and \ref{con:stability}. 
\end{theorem}

To ensure that the network is stable at all stages, we pre-train the feedback weights with a large $\alpha$ to guarantee network stability while aligning the feedback weights with Eq. \eqref{eq:fb_alignment}, after which we start the full network training with a small $\alpha$. Theorem \ref{theorem:fb_learning} considers training the feedback weights on a single input sample. In reality, however, multiple input samples are used for training. For linear networks, the network Jacobian $\Js$ is independent of the input samples and Theorem \ref{theorem:fb_learning} holds exactly. For nonlinear networks, $\Js$ will change depending on the input, hence, the feedback plasticity \eqref{eq:Q_dynamics_scaled} will drive $Q$ to align with some average of Eq. \eqref{eq:fb_alignment} for all input samples. 

\paragraph{Debiasing the forward weight updates.}
While we need noise correlations for training the feedback weights, these can bias the forward weight updates (see App. \ref{app:noise_bias}). We solve this challenge by low-pass filtering the pre-synaptic plasticity signal for the forward weight updates, such that the pre-synaptic noise fluctuations are removed:
\begin{align}
\label{eq:debiased-dW}
    \tau_W \ddt W_i(t) &= \Big(\phi\big(\vv_i(t)\big) - \phi\big(W_i\vr_{i-1}(t)\big)\Big) \vr_{\mathrm{lp}, i-1}^T,
\end{align}
with $\vr_{\mathrm{lp,} i} \triangleq \frac{1}{\tau_f}\int_0^t \exp\big({-}\frac{1}{\tau_f}(t-t')\big)\vr_{i}(t')\dd t'$ an exponential moving average of $\vr_i$ with filtering time constant $\tau_f$.

\section{Experiments}
\label{experiments}

Here, we validate our theoretical findings with empirical results. By studying in detail a synthetic toy example and modest computer-vision benchmarks, we show that the minimizing control framework succeeds in minimizing the training loss $\calL$, and that Strong-DFC equipped with single-phase learning succeeds in approximating the gradient of $\calH$, thereby validating itself as a principled CA method. Finally, we show that Strong-DFC is significantly more noise-robust compared to standard DFC, due to its strong feedback influences on neural activity.\footnote{Source code for all experiments is available at: \texttt{\url{https://github.com/mariacer/strong_dfc}}.}

\paragraph{Strong-DFC performs CA by minimizing feedback control.}

We start with validating our main theoretical claim that minimizing control is a suitable new theoretical framework for learning. To this end, we first focus on a low-dimensional teacher-student nonlinear function approximation problem, which can be efficiently simulated.

In this problem, the goal of learning is to reduce the squared error $\mathcal{L} = \mathbb{E}_{\mathbf{x} \sim \mathcal{N}(0, I)} [\|f_\text{trgt}(\mathbf{x}) -  \mathbf{v}_{L,\mathrm{ss}}(\mathbf{x})\|^2 ]$ between the target $f_\text{trgt}(\mathbf{x})$ provided by the teacher network we wish to approximate and the steady-state last-layer activity of our student network, which is a function of the input $\ve{x}$. 
The student network is trained with Strong-DFC, and its input layer activity $\ve{r}_0$ is clamped to the same random pattern $\mathbf{x} \sim \mathcal{N}(0, I)$ used to probe the teacher function. To confirm that Theorem~\ref{theorem:grad_H} and Proposition \ref{proposition:min_H_min_L} hold in practice, we study an idealized setting. We disable neural noise ($\sigma = 0$), simulate the neural dynamics \eqref{eq:noisy_dynamics_v} to equilibrium, and then apply a single weight change to $W$ according to the steady-state Strong-DFC update \eqref{eq:update_W_ss}. Moreover, to ensure that Condition \ref{con:col_space} is satisfied in this idealized setting at all points during training, we manually set the feedback weights equal to the transposed network Jacobian: $Q = \left.J^T\right\vert_{\vv = \vv(t)}$ (see Condition \ref{con:col_space}). We use Euler's forward method to simulate the dynamics and we initialize the network state to the feedforward prediction for each datapoint.

\begin{figure}[t]
     \centering
    \includegraphics[width=0.47\textwidth]{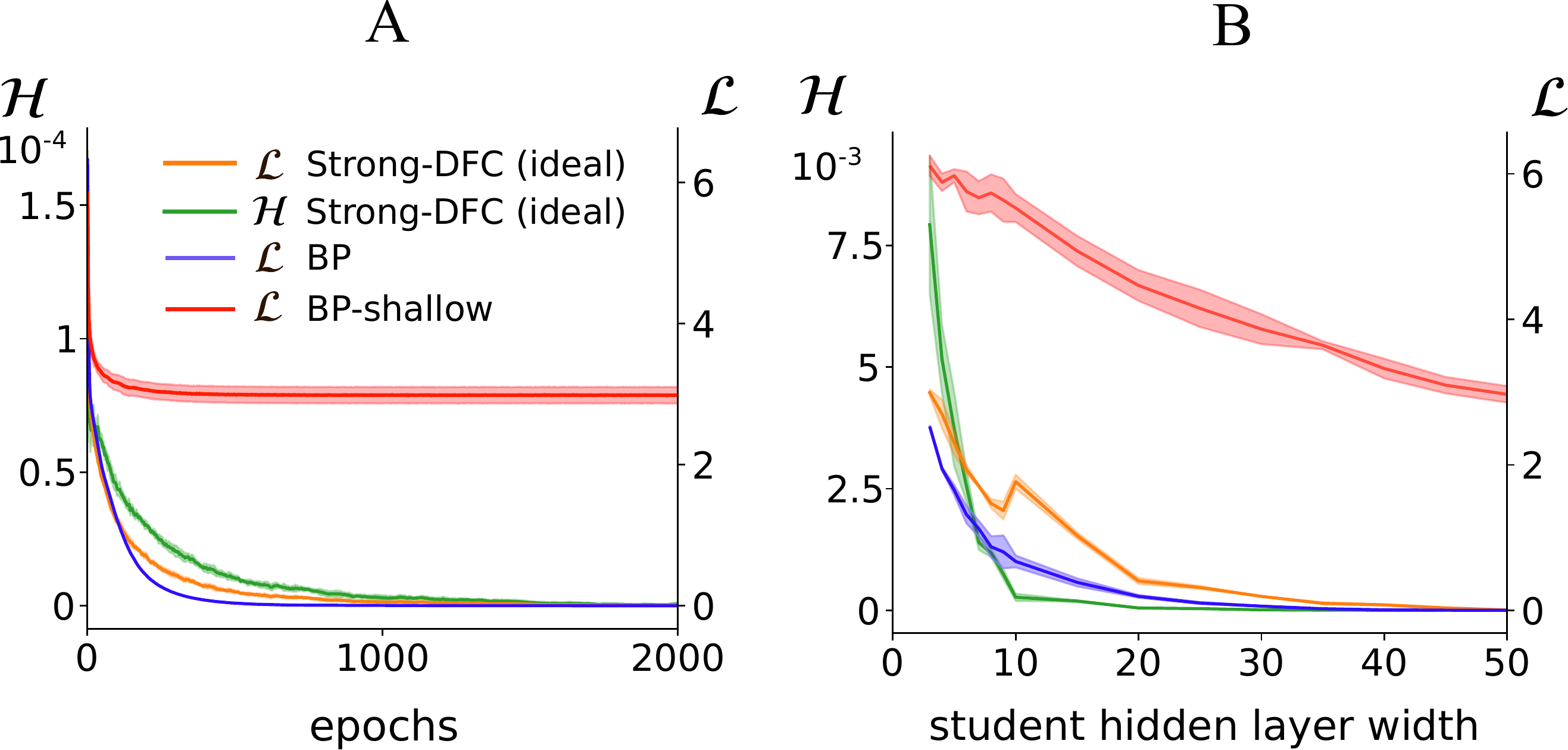}
     \caption{\textbf{Strong-DFC minimizes the original training loss and the amount of control.} (A) The training loss $\mathcal{L}$ as well as the surrogate loss $\mathcal{H}$ indicating the amount of control received are minimized during training with Strong-DFC and ideal feedback weights. The results are obtained in a nonlinear student-teacher setting with a fixed teacher of size 30-10-10-10-5, and an overparameterized student of size 30-50-50-50-5, both equipped with $\tanh$ nonlinearities. (B) As hinted by Proposition \ref{proposition:min_H_min_L}, the amount of control $\mathcal{H}$ at the end of training depends on how well $\mathcal{L}$ can be minimized, and is thus dependent on the capacity of the network. Same setting as in (A), with the loss at convergence plotted in function of the width of the hidden layers of the student that varies from 3 to 50. We compare the Strong-DFC results against backpropagation (BP) and a student network of the same architecture but where only the last layer is trained (BP-shallow).}
    \label{fig:overparam}
      
\end{figure}

Our experiments confirm that the surrogate loss $\mathcal{H}$ is reduced as training progresses, eventually reaching a low value (of the order of $10^{-7}$), in accordance with Theorem~\ref{theorem:grad_H} (see Fig. \ref{fig:overparam}A). Crucially, we also observe that as $\mathcal{H}$ decreases, so does the actual loss of interest $\mathcal{L}$. At the end of training, $\mathcal{L}$ is of the order $10^{-3}$, which supports our claim that the minima of the two loss functions coincide (cf. Proposition~\ref{proposition:min_H_min_L}). Furthermore, by varying the width of the hidden layers of the student (Fig. \ref{fig:overparam}B), we show that the amount of control $\mathcal{H}$ required at the end of training depends on how well the original training loss $\mathcal{L}$ can be minimized, with more overparameterized networks reaching lower loss $\mathcal{L}$ and requiring less help $\mathcal{H}$.
We note that such low $\mathcal{L}$ cannot be achieved by a linear function approximator, nor by a shallow student that only learns its output weights. This confirms that the ideal version of Strong-DFC performs useful credit assignment. We conclude that learning by minimizing control succeeds in solving our nonlinear function approximation problem.

We confirm the above results for the idealized setting on the MNIST dataset \citep{lecun_mnist_2010} (Table \ref{tab:idealQ}). Here we also observe that both the original training loss $\mathcal{L}$ and the surrogate loss $\mathcal{H}$ reach low values, which in turn translates into high testing accuracy. 
Overall, these results confirm that the ideal version of Strong-DFC performs useful credit assignment in more challenging problems. 
\begin{table}[h]
    \centering
        \caption{Strong-DFC (with ideal feedback weights) successfully minimizes the original loss $\mathcal{L}$ and the surrogate loss $\mathcal{H}$ in MNIST. Results (mean $\pm$ std, 5 seeds) using a multi-layer perceptron with three hidden layers of 256 neurons. Values are reported for the epoch with lowest training loss out of 40.}
    \label{tab:idealQ}
\begin{tabular}{ c  c  c }
\toprule
& Strong-DFC (ideal) & BP \\ 
\midrule
\midrule
\multicolumn{1}{l}{Train loss} $\mathcal{L}$ & $8.09^{\pm 0.35} \cdot 10^{-2}$ & $3.38^{\pm 1.30} \cdot 10^{-5}$ \\
\multicolumn{1}{l}{Test loss} $\mathcal{L}$ & $1.71^{\pm 0.10} \cdot 10^{-1}$ & $9.60^{\pm 0.04}\cdot 10^{-2}$ \\
\multicolumn{1}{l}{Train error} & $0.07^{\pm 0.09}\%$ & $0.00^{\pm 0.00}\%$ \\
\multicolumn{1}{l}{Test error} & $1.98^{\pm 0.11}\%$ & $1.78^{\pm 0.04}\%$ \\
\multicolumn{1}{l}{Train loss} $\mathcal{H}$ & $2.64^{\pm 3.29} \cdot 10^{-5}$ & N/A \\
\end{tabular}
\end{table}

\paragraph{Feedback learning dynamically tracks feedforward learning.}

\begin{figure*}[h]
\centering
\includegraphics[width=0.78\textwidth]{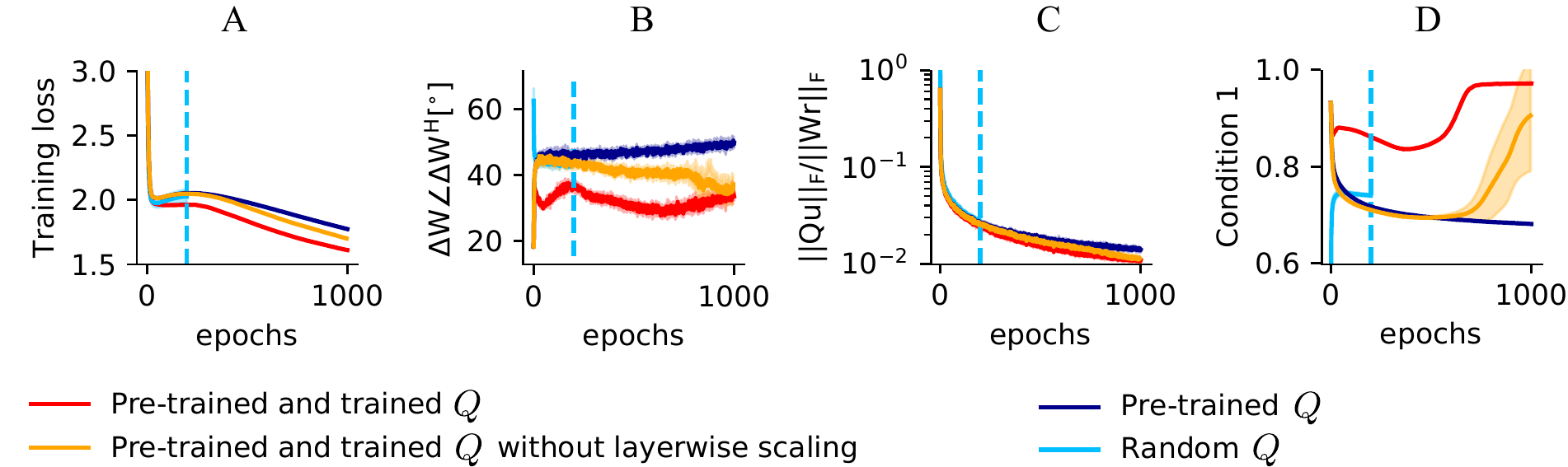}
\caption{\textbf{Strong-DFC simulations for a nonlinear student-teacher regression task}. Standard Strong-DFC, where the feedback weights $Q$ are both pre-trained before training the forward weights and trained simultaneously with the forward weights according to Eq. \ref{eq:Q_dynamics_scaled} (red), is compared to a set of controls where the feedback weights are: trained without the $(1 + \tau_v / \taue)^{(L-i)}$ scaling term in the updates (orange), only pre-trained (dark blue), or kept fixed to a random initialization during the entire simulation (light blue). The blue vertical dotted lines indicate the moment where training with random feedback weights leads to instabilities.(A) Strong-DFC successfully reduces the loss $\mathcal{L}$ during training.
(B) When simultaneously learning the feedback weights, the forward weight updates of Strong-DFC align well with the ideal updates following the gradient of $\mathcal{H}$, showing that the amount of control is being minimized with our learning rules. (C) This is achieved through the use of \emph{strong} feedback, which allows driving the network to the desired output even at early stages of training when the ratio of feedback ($\|Qu\|_2^2$) to forward ($\|Wr\|_2^2$) input is high. (D) The single-phase learning scheme succeeds in driving the feedback weights to satisfying Condition \ref{con:col_space}, enabling efficient training of the forward weights. We measure the ratio between the norm of Q projected into the row space of $\Js$, and $Q$ itself, with a value of $1.0$ representing perfect compliance (see App. \ref{app:experiments}).  The results are obtained in a student-teacher setting with a fixed teacher of size 30-10-10-10-5, and an overparameterized student of size 30-50-50-50-5 with $\tanh$ nonlinearities. Shaded regions indicate standard deviations across five runs.}
\label{fig:toy:curves}
\end{figure*}

Having verified that Strong-DFC fulfills the minimizing control objective, and that this solves our task, we move to a more challenging setting where feedback weights are no longer manually fixed to ideal values, and where neural activity is noisy. The learning problem is otherwise unchanged. In these experiments, we now adjust the feedback weights $Q$ according to our anti-Hebbian plasticity rule \eqref{eq:Q_dynamics_scaled} concurrently as the forward weights evolve through our always-on, debiased plasticity rule \eqref{eq:debiased-dW}. For computational efficiency, we accumulate our weight changes over time and apply them once the neural steady-state is reached. We compare Strong-DFC to various ablations on the feedback weight learning (Fig. \ref{fig:toy:curves}), and focus on the initial stage of learning as this provides the clearest differences. 
We observe that Strong-DFC successfully decreases the training loss (Fig. \ref{fig:toy:curves}A) through updates that approximately align with the negative gradient of $\mathcal{H}$ (Fig. \ref{fig:toy:curves}B). Furthermore, our single-phase feedback learning rule succeeds in dynamically tracking the ongoing feedforward learning process and in finding values of $Q$ which approximately respect Condition~\ref{con:col_space} (Fig. \ref{fig:toy:curves}D). We note that throughout our simulation the feedforward and feedback inputs to any given neuron are of the same order of magnitude during early training, which confirms that the network operates in the strong feedback regime (Fig. \ref{fig:toy:curves}C). Importantly, training performance, alignment with the gradient of $\mathcal{H}$ and compliance with Condition~\ref{con:col_space} are all negatively impacted whenever the feedback weights $Q$ are not trained during the simulation. Furthermore, without any pre-training, updates quickly lead to numerical instabilities. This indicates that purely random feedback mappings are not sufficient for activity-dependent learning with strong feedback modulation, in contrast to random feedback mappings that are solely used for plasticity \citep{lillicrap2016random, nokland2016direct}.

\paragraph{Strong-DFC exhibits robustness to noise and competitive performance in standard computer vision benchmarks.}
We now investigate if Strong-DFC can successfully perform credit assignment on a deeper neural network model, that we apply to two standard image classification benchmarks, MNIST and Fashion-MNIST \citep{xiao_fashion-mnist_2017}. Moreover, we set the focus of our investigation on noisy neural states, where $\sigma$ in \eqref{eq:noisy_dynamics_v} is non-negligible ($\sigma \sim 10^{-2}$), to determine whether the strong feedback introduced in Strong-DFC enables learning in the high-noise regime, and whether our plasticity rule \eqref{eq:Q_dynamics_scaled} can use the fluctuations to learn feedback connections that are useful for credit assignment on deeper networks. In this section, we study a fully-connected three-hidden-layer neural network (256 units per layer) with a final softmax classification layer and choose the cross-entropy as our loss function $\mathcal{L}$, which can be handled in our framework as detailed in Appendix \ref{app:softmax}. When using Strong-DFC, we learn feedback and feedforward weights concurrently with always-on plasticity rules; plasticity is only interrupted when switching patterns and re-initializing the network states. To ensure network stability at the beginning of training, we pre-train the feedback weights using Eq. \eqref{eq:Q_dynamics_scaled} and a high controller leak rate $\alpha$. 

\begin{table}[h]
    \centering
        \caption{Computer vision results. Test errors (mean $\pm$ std, 5 seeds) obtained after 40 training epochs on a multi-layer perceptron with three hidden layers of 256 neurons each. For DFC, results are provided without noise (original) and with noise during the forward weight training. For Strong-DFC, we consider the (original) single-phase setting with noise and a two-phase setting without noise in the forward weights training phase.}
    \label{tab:my_label2}
\begin{tabular}{c@{\hskip 1pt}c@{\hskip 2pt}c}
\toprule
MNIST- & standard ($\%$) & fashion ($\%$)\\
\midrule
\midrule
\multicolumn{1}{l}{BP} & $1.83^{\pm 0.11}$ & $10.60^{\pm 0.44}$\\ 
\midrule
\multicolumn{1}{l}{Strong-DFC} (with noise) & $2.19^{\pm 0.05}$ & $12.07^{\pm 0.16}$\\
\multicolumn{1}{l}{DFC (with noise)} & $15.15^{\pm 0.44}$ & $16.29^{\pm 0.41}$\\
\midrule
\multicolumn{1}{l}{Strong-DFC (no noise)} & $1.98^{\pm 0.05}$ & $11.36^{\pm 0.17}$\\
\multicolumn{1}{l}{DFC (no noise)}  & $2.09^{\pm 0.10}$ & $11.31^{\pm 0.14}$\\

\end{tabular}
\end{table}

In Table~\ref{tab:my_label2}, we compare the test set prediction error rates of Strong-DFC to DFC, as well as to standard backpropagation (BP), which we provide as a reference baseline. We optimize the hyperparameters of each method independently, for best performance on a validation set of 5000 datapoints. We use BP to train a noiseless network, and DFC to train both a noiseless and a noisy network, where we add noise to the dynamics according to  Eq. \eqref{eq:noisy_dynamics_v}. In the noisy setting, we equip DFC with the debiased plasticity rule \eqref{eq:debiased-dW} for fair comparison. As previously reported \citep{meulemans2021credit}, in the noiseless case, the test-set performance of DFC is close to that achieved by BP. However, the situation changes in the large-$\sigma$ regime: the performance of (weak feedback) DFC drastically drops, whereas Strong-DFC is undisturbed on MNIST and not severely impaired on Fashion-MNIST (we further illustrate this in a student-teacher regression setting in Appendix \ref{app:noise_robustness}). When we train the forward weights with Strong-DFC in a separate phase without noise, its performance is competitive to both DFC and BP. Moreover, we observed that without feedback plasticity, we did not succeed in training our network when starting from a random initial $Q$, as the network was unstable. Our feedback plasticity thus succeeds in pushing the initial network state to an appropriate one and in keeping it throughout learning.

\section{Discussion}

Modern deep neural networks are invariably learned by backpropagation-of-error. While backpropagation has proven to be a highly effective gradient-based credit assignment method for training deep artificial neural networks, it is unsatisfactory as a model of learning in cortical networks in at least three important aspects. First, it needs precisely-symmetric forward and feedback connections. Second, backpropagated errors do not influence neural activity. Third, it requires two precisely-clocked phases. Instead of attempting to remedy these issues one by one, we propose a fundamentally new approach for deep learning, that overcomes all of these by design, and provide first results indicating that it is an effective method for credit assignment.

Our study shows that these three critical issues can be overcome at once within the framework of gradient-based optimization, by casting learning as a control minimization problem. For this, we first augment standard deep neural networks with a feedback controller, which dynamically changes the hidden neural activity to reach the desired output. Then, we derive a synaptic plasticity rule with the objective of minimizing the influence of the feedback controller on the neural activity. This principled derivation led us to Strong-DFC, a variation of the recently proposed DFC method which inherits all of its appealing properties as a biologically plausible deep learning framework. Like DFC, our weight update rule is local in space and time for a broad range of (not necessarily symmetric) feedback connectivity patterns and neural feedback controller architectures, and it can be interpreted as a biologically inspired dendritic voltage-dependent synaptic plasticity rule \citep{urbanczik2014learning}. Importantly, however, top-down feedback signals are no longer required to be vanishingly small for Strong-DFC.

This strong feedback influence sets our minimizing control framework apart from previous theories of how the brain might estimate objective function gradients by measuring small changes in neural activity, generated by `nudging' the network towards a slightly better state \citep{hinton1988learning, o1996biologically, xie2003equivalence, scellier2017equilibrium, sacramento2018dendritic,  lillicrap2020backpropagation, meulemans2021credit}. In biology, such theories might face a fundamental obstacle during learning, since cortical activity is notoriously noisy \citep{rusakov2020noisy}, making it hard to estimate small activity changes as plasticity signals. Moreover, they are at odds with a growing number of experimental neuroscience observations reporting large neural activity changes in the presence of unexpected or novel events, which are thought to drive synaptic plasticity \citep{keller2018predictive}, and the existence of top-down feedback signals that are sufficient to drive secondary receptive field responses in pyramidal neurons \citep{keller2020feedback}. 
Our minimizing control framework shows how to perform principled mathematical optimization of an objective function, while being consistent with these findings.

A long-standing question in neuroscience is whether noise plays a computational role in the brain -- whether it is `a bug or a feature' of biological neural circuits \citep{rusakov2020noisy}. Our study suggests that noise could play a key role in ensuring that gradient-based credit assignment information is transmitted backwards through the network. Building on the noise-tolerance of Strong-DFC, we leverage noise to concurrently and seamlessly learn both feedforward and feedback weights so that Condition \ref{con:col_space} is constantly fulfilled. In contrast to standard two-phase learning of forward and backward connections \citep{akrout2019deep, lee2015difference,  meulemans2021credit}, our plasticity model does not require any pauses, distinct phases, plasticity switches, or alternations between noise- and input-driven modes. Our always-on plasticity rules are arguably more biologically plausible than alternative two-phase learning algorithms, and simpler to implement in neuromorphic hardware.

In our pyramidal cell model, top-down apical synapses transmit feedback information which instructs the plasticity of bottom-up basal synapses, an idea that can be traced back to the seminal work of \citet{kording2001supervised}. Interpreted as such, our results suggest that the plasticity rules for top-down and bottom-up connections might display major differences, which is experimentally testable. Both rules include a voltage-dependent postsynaptic factor, in agreement with phenomenological plasticity models \citep{clopath2010voltage}. However, the presynaptic components of our apical and basal plasticity rules are high- and low-pass filtered, respectively. Such presynaptic activity filters are a recurring element in reward-driven synaptic plasticity rules \citep{seung2003learning}. Moreover, our apical plasticity rule is anti-Hebbian and its postsynaptic term depends only on the local dendritic voltage, not on the somatic spiking activity. This may be consistent with the fact that, for pyramidal cells, backpropagating action potentials often fail to reach the apical dendrite, questioning their role as the main drivers of apical synaptic plasticity \citep{spruston2008pyramidal, gambino2014sensory}.

We highlight that, despite the close methodological connections between DFC and Strong-DFC, our minimizing control framework is fundamentally different from the underlying theory of DFC and many other recent biologically plausible learning methods that link their weight updates to a new variant of Gauss-Newton optimization with a minibatch size of one \citep{meulemans2021credit, meulemans2020theoretical, podlaski2020biological, bengio2020deriving}. While the specific optimization characteristics of these methods are not yet fully understood, our minimizing control framework can be cast as gradient descent on a surrogate loss function. Hence, the rich field studying the application of stochastic gradient descent to neural networks \citep{bottou2018optimization} can be applied on our new loss to uncover the optimization characteristics of Strong-DFC. Furthermore, our minimizing control approach draws from control theory, but it is distinct from standard optimal control. In particular, it is well known that the adjoint state method from optimal control is equivalent to backpropagation on the supervised loss function, when applied to deep feedforward neural networks \citep{lecun1988theoretical}, and hence different from our minimizing control framework that performs gradient descent on a surrogate loss function.

In practice, the Strong-DFC updates do not exactly follow the negative gradient of $\calH$, due to (i) the limited number of iterations for training the feedback weights, (ii) the limited capacity of the linear pathway $Q$ to satisfy Condition \ref{con:col_space} for every input sample, and (iii) the difference between the steady-state weight update \eqref{eq:update_W_ss} and the Strong-DFC weight update \eqref{eq:W_dynamics}. Despite these challenges, Fig \ref{fig:toy:curves} shows that Strong-DFC approximates the theory well. Future work can further improve Strong-DFC by investigating alternative feedback mappings and network architectures. As Strong-DFC uses continuous dynamics, it is costly to simulate on conventional deep learning hardware, preventing it from being tested on large-scale machine learning benchmarks. A promising solution is to implement Strong-DFC on analog and neuromorphic hardware, where the dynamics can coincide with the physical dynamics of the analog components. Consequently, this can make Strong-DFC an attractive principled approach for CA on analog deep learning implementations, commonly used in Edge-AI and other low-energy applications of deep learning \citep{xiao2020analog, misra2010artificial}.


By providing a novel optimization approach for deep neural networks that differs from the usual direct minimization of the output loss, our minimizing control framework allowed us to derive a novel gradient-based credit assignment method that relies on strong feedback influence. Instantiating our framework with Strong-DFC naturally leads to biologically desired characteristics, such as single-phase learning, flexible feedback connectivity requirements, and local learning rules, further underlining its promise for understanding credit assignment in the brain.

\section*{Acknowledgements}
This work was supported by the Swiss National Science Foundation (B.F.G. CRSII5-173721 and 315230\_189251), ETH project funding (B.F.G. ETH-20 19-01) and the Human Frontiers Science Program (RGY0072/2019). João Sacramento was supported by an Ambizione grant (PZ00P3\_186027) from the Swiss National Science Foundation. We would like to thank Nicolas Zucchet for his valuable feedback and engaging discussions on the optimization theory aspects of the minimizing control framework; and Aditya Gilra for insightful discussions on the learning through control framework.


\bibliography{biblio}
\bibliographystyle{icml2022}

\newpage
\appendix

\subfile{appendices/appendix}

\end{document}

%% file: appendices/appendix.tex
\onecolumn

\setcounter{figure}{0} \renewcommand{\thefigure}{S\arabic{figure}}
\setcounter{theorem}{0} \renewcommand{\thetheorem}{S\arabic{theorem}}
\setcounter{condition}{0} \renewcommand{\thecondition}{S\arabic{condition}}
\setcounter{table}{0} \renewcommand{\thetable}{S\arabic{table}}



\section{Proofs and discussion for Section \ref{sec:minimizing_control}}
\subsection{Proof Theorem 1}\label{app:proof1}
In this appendix section, we prove Theorem \ref{theorem:grad_H}, which we copy below for convenience.
\begin{theorem}\label{theorem_app:grad_H}
Assuming $\Js Q + \talpha I$ is invertible, the total derivative of $\calH$ w.r.t. $W_i$ is given by:
\begin{align}\label{eq_app:total_grad_H}
    \der{W_i}{\calH} = -\sum_b J_{i, \sss}^{(b)T} \left(\Js^{(b)} Q + \talpha I\right)^{-T}Q^TQ  \vus^{(b)} \vr_{i-1, \sss}^{(b)T},
\end{align}
with $J_{i, \sss} \triangleq \pder{\vv_i}{\vrl}$ and $J_{\mathrm{ss}} \triangleq [\frac{\partial \vr_L}{\partial \vv_1}, ...,\frac{\partial \vr_L}{\partial \vv_L}]$, both evaluated at steady state, and $\talpha \triangleq \frac{\alpha}{1 + k \alpha}$.

Furthermore, assuming Condition \ref{con:col_space} holds, the network is stable, $\alpha \rightarrow 0$, and $\J_{\mathrm{ss}}Q + \talpha I$ is invertible, then the following steady-state (ss) updates for the forward weights, averaged over the batch samples,
\begin{align}\label{eq_app:update_W_ss}
    \Delta W_{i, \mathrm{ss}} = \eta (\ve{v}_{i,\mathrm{ss}} - \ve{v}_{i,\mathrm{ss}}^{\mathrm{ff}})\ve{r}_{i-1,\mathrm{ss}}^T\,\,,
\end{align}

\vspace{-0.3cm}
with $\eta$ the stepsize, align with the negative gradient $-\frac{\dd \mathcal{H}}{\dd W_i}$.
\end{theorem}

\begin{condition}\label{con_app:col_space}
The column space of $Q$ is equal to the row space of $J_{\mathrm{ss}} \triangleq [\frac{\partial \vr_L}{\partial \vv_1}, ...,\frac{\partial \vr_L}{\partial \vv_L}]$ at steady state.
\end{condition}

The intuition behind the proof is as follows. First, we compute in Lemma \ref{lemma_app:grad_H} the gradient of $\mathcal{H}$ using the Implicit Function Theorem, as this provides us with the means to calculate how the control signal $\vu_{\sss}$ at steady steady state changes when $W_i$ is changed. In Lemma \ref{lemma_app:grad_H_local}, we show that $\frac{\dd \mathcal{H}}{\dd W_i}$ collapses to a local update if Condition \ref{con:col_space} is satisfied. Finally, we prove Theorem \ref{theorem:grad_H} by showing that this local update is equal to the steady-state weight update in Eq. \eqref{eq:update_W_ss}. 

\begin{lemma}
\label{lemma_app:grad_H}
The gradient of $\mathcal{H}$ w.r.t. $\vt$ for a single sample, evaluated at steady state, is given by 
\begin{align}
\frac{1}{2}\der{\vt}{\|Q\vus\|^2_2} = - \vu_{\sss}^TQ^TQ\left(J_{\sss}Q + \tilde{\alpha} I \right)^{-1}J_{\sss}R_{\sss}^T,
\end{align}
with $\vt = [\vec{W}_1^T, ... \vec{W}_L^T]^T$ and $R_{\sss}$ defined below. 
\begin{align} \label{eq_app:Rss_matrix}
    R_{\sss}^T \triangleq \begin{bmatrix} 
    \vr_0^T \otimes I & 0 & 0 \\
    0 & \ddots & 0 \\
    0 & 0 & \vr_{L-1}^T \otimes I
    \end{bmatrix}
\end{align}
with $\otimes$ the Kronecker product.
\end{lemma}
\begin{proof}
Using the chain rule, we get the following expression for the gradient of interest: 
\begin{align}
    \frac{1}{2}\der{\vt}{\|Q\vus\|^2_2} = \vus^TQ^TQ \der{\vt}{\vus}
\end{align}
Now we proceed to find $\der{\vt}{\vus}$. At steady state, the network and controller dynamics result in the following equilibrium equations:
\begin{align}
    \vv_{i, \sss} &= W_i \phi(\vv_{i-1, \sss}) + Q_i \vu_{\sss}, \quad 1 \leq i \leq L \label{eq_app:network_ss}\\
    \vu_{\sss} &= k\ve{e}_{\sss} + \uint_{\sss} \label{eq_app:controller_ss}\\
    0 &= \ve{e}_{\sss} - \alpha \uint_{\sss} \\
    \ve{e}_{\sss} &= \vrl^* - \vr_{L,\sss}
\end{align}
The equilibrium solutions in Eq. \eqref{eq_app:network_ss} can be used to obtain an explicit function of $\vrl$ w.r.t. $\vt$ and $\vu_{\sss}$, which in turn can be used to obtain an explicit function of $\ve{e}_{\sss}$ i.f.o. $\vt$ and $\vu_{\sss}$. Combining all equations results in the following implicit function relating $\vu_{\sss}$ to $\vt$:
\begin{align} \label{eq_app:controller_steady_state}
    F(\vu_{\sss}, \vt) \triangleq \ve{e}_{\sss}(\vu_{\sss}, \vt) - \tilde{\alpha} \vus = 0,
\end{align}
with $\talpha \triangleq \frac{\alpha}{1+ \alpha k}$. If $\pder{\vus}{F}$ is invertible, we can use the Implicit Function Theorem to obtain the total derivative of $\vus$ w.r.t. $\vt$:
\begin{align}
    \der{\vt}{\vus} = - \left( \pder{\vus}{F}\right)^{-1} \pder{\vt}{F}.
\end{align}
Investigating each part separately, we get
\begin{align}\label{eq_app:jacobian_steady_state_controller}
    \pder{\vus}{F} &= - \pder{\vu}{\vrl} - \tilde{\alpha} I \\
    &= -\sum_{i=1}^L \pder{\vv_i}{\vrl} \pder{\vu}{\vv_i} - \tilde{\alpha} I \\ 
    &= -J_{\sss} Q - \tilde{\alpha}I,
\end{align}
with all derivatives evaluated at the steady-state activations, and
\begin{align}
    \pder{\vt}{F} &= \pder{\vt}{\vrl} \\
    &= - J_{\sss} R_{\sss}^T,
\end{align}
with $R_{\sss}^T$ defined in Eq. \eqref{eq_app:Rss_matrix}. Bringing everything together results in 
\begin{align}
    \der{\vt}{\vus} &= -\left(J_{\sss} Q + \talpha I \right)^{-1} \Js \Rs^T \\
    \frac{1}{2}\der{\vt}{\|Q\vus\|^2_2} &= -\vus^TQ^TQ \left( J_{\sss} Q + \talpha I \right)^{-1}  \Js \Rs^T.
\end{align}
\end{proof}

\begin{lemma}\label{lemma_app:grad_H_local}
When Condition \ref{con:col_space} is satisfied and $\Js Q$ is invertible, $\der{W_i}{\|Q\vus \|^2_2}$ results in a local update rule in the limit of $\talpha \rightarrow 0$: 
\begin{align}
    \lim_{\talpha \rightarrow 0} \frac{1}{2}\der{W_i}{\|Q\vus\|^2_2} = -Q_i\vus \vr_{i-1, \sss}^T.
\end{align}
\end{lemma}
\begin{proof}
Using a similar proof as Lemma S2 in \citet{meulemans2021credit}, we have that
\begin{align}
    \lim_{\talpha \rightarrow 0} \left(J_{\sss} Q + \talpha I \right)^{-1}\Js = Q^{\dagger}
\end{align}
iff $\mathrm{Row}( \Js) = \mathrm{Col}(Q)$, with $Q^{\dagger}$ the Moore-Penrose pseudoinverse of $Q$. Using the fact that $Q^TQQ^{\dagger} = Q^T$, we have that 
\begin{align}
    \lim_{\talpha \rightarrow 0}\frac{1}{2}\der{\vt}{\|Q\vus\|^2_2} &= -\vus^TQ^T\Rs^T
\end{align}
iff $\mathrm{Row}(\Js) = \mathrm{Col}(Q)$ and $\Js$ is of full row rank. Going from the vectorized notation for $\vt$ towards the matrix notation for $W_i$ gives us 
\begin{align}
     \lim_{\talpha \rightarrow 0} \frac{1}{2}\der{W_i^T}{\|Q\vus\|^2_2} = -Q\vus \vr_{i-1, \sss}^T.
\end{align}
\end{proof}

Now, we are ready to prove Theorem \ref{theorem:grad_H}. 
\begin{proof}
From the linearity of the derivative operator, we have that
\begin{align}
    \der{W_i^T}{\calH} = \sum_b \frac{1}{2}\der{W_i}{\|Q\vus^{(b)}\|^2_2}.
\end{align}
Rewriting Lemma \ref{lemma_app:grad_H} from its vectorized form towards matrix form, and filling it in the above equation, proves the first part of Theorem \ref{theorem:grad_H}: 
\begin{align}
    \der{W_i^T}{\calH} = -\sum_b J_{i, \sss}^{(b)T} \left(\Js^{(b)} Q + \talpha I\right)^{-T}Q^TQ  \vus^{(b)} \vr_{i-1, \sss}^{(b)T},
\end{align}

For the second part of the Theorem, the conditions of Lemma \ref{lemma_app:grad_H_local} are satisfied by the stated assumptions of this theorem. Hence, we have that:
\begin{align}
    \lim_{\talpha \rightarrow 0}\der{W_i^T}{\calH} = -\sum_b Q_i\vus^{(b)} \vr_{i-1, \sss}^{(b)T}.
\end{align}
Finally, from the steady-state solution for the network dynamics, we have that $\vv_{i,\sss} - \vfb_{i,\sss} = Q_i \vus$, thereby concluding the proof.
\end{proof}

\subsection{Interpreting Condition \ref{con:col_space}} \label{app:col_condition}
Here, we investigate in more depth why Condition \ref{con:col_space} is needed in Theorem \ref{theorem:grad_H}. The main line of argumentation goes as follows: if Condition \ref{con:col_space} is not satisfied, the steady-state weight update given by Eq. \eqref{eq:update_W_ss} has components inside the nullspace of the network, and hence, can only partially influence the network output. The gradient of $\|Q\vus\|_2^2$ is its steepest ascent direction, thus all the components of this gradient should influence the network output and consequently, the amount of needed control feedback, because otherwise there would exist a 'steeper' direction. Therefore, if the weight update given by Eq. \eqref{eq:update_W_ss} contains nullspace components, it cannot be fully aligned with the gradient.  

Now, let us specify the \textit{nullspace} of the network. As the dimension of the concatenated forward weights $\vt$ is higher than the dimension of the network output, there exist certain weight updates $\Delta \vt$ that do not result in any change to the output $\vrl$ of the network, as they lie inside the nullspace of the network. For small weight updates, the nullspace of the network at steady state is connected to the network Jacobian, as we have that $\Delta \vrl \approx \Js \Rs^T \Delta \vt$, with $\Rs$ as defined in Lemma \ref{lemma_app:grad_H}. From this formulation, one can see that the nullspace of the network is equal to the nullspace of $ \Js \Rs^T$. 

When vectorizing the steady-state weight updates (Eq. \ref{eq:update_W_ss}), we see that they lie inside the rowspace of $Q^T \Rs^T$. If we have that the column space of $Q$ (hence, the rowspace of $Q^T$) is equal to the row space of $\Js$, the steady-state weight updates will always lie inside the rowspace of $\Js \Rs^T$, and hence, can never lie inside the nullspace of the network. 

The negative gradient of $\|Q\vus\|_2^2$ is its steepest descent direction, thus all the components of the gradient should influence the network output and consequently the amount of needed control input. Therefore, if the weight update Eq. \eqref{eq:update_W_ss} contains nullspace components, it cannot be fully aligned with the gradient. Consequently, Condition \ref{con:col_space} is a necessary condition for the steady-state update (Eq. \ref{eq:update_W_ss}) to align with the negative gradient. The sufficiency of Condition \ref{con:col_space} follows from the proof of Theorem \ref{theorem:grad_H}.

To illustrate the above arguments further, let us investigate in more detail what happens when we update the forward network parameters $\vt$. Assuming a small weight update $\Delta \vt$, we can do a first-order Taylor expansion of the steady-state network output after the weight update:
\begin{align}
    \vr_{L,\sss}^{(m+1)} \approx \vr_{L,\sss}^{(m)} + J_{\vt}\Delta \vt + \Js Q \Delta \vus,
\end{align}
with $J_{\vt} = \pder{\vt}{\vrl} = \Js \Rs^T$ and $m$ the training iteration. The weight update changes the amount of control needed to reach the output target, which we capture by introducing $\Delta \vus$. As the feedback controller drives the network to the supervised label during each training iteration (assuming $\alpha \rightarrow 0$), we have that $\vr_{L,\sss}^{(m+1)} = \vr_{L,\sss}^{(m)} = \vrltrue$ and hence
\begin{align}
    \Js \Rs^T \Delta \vt \approx - \Js Q \Delta \vus.
\end{align}
The above equation relates the update $\Delta \vt$ to the change in steady-state control input $Q\Delta \vus$. In order to maximally reduce $\calH$ and hence $\snorm{Q\vus}^2_2$, the magnitude of the LHS should be as big as possible. As a result, we want $\Delta \vt$ to lie fully inside the rowspace of $\Js \Rs^T$, because otherwise, $\Delta \vt$ will partially be lost in its nullspace, reducing the magnitude of the LHS. As $\Delta \vt = \Rs Q \vus$, we have that the columnspace of $\Rs Q$ needs to lie inside the rowspace of $\Js \Rs^T$. This condition is equivalent to the columnspace of $Q$ lying inside the rowspace of $\Js$. Note that the above arguments should be interpreted on an intuitive level, a detailed mathematical consideration of the above arguments is more nuanced, but leads to the same high-level intuition.

\subsection{Effect of the nonlinearity $\phi$ on the weight updates}\label{app:nonlinearity_update}
The main effect of including the nonlinearity $\phi$ into the Strong-DFC weight updates (Eq. \ref{eq:W_dynamics}), in contrast to the steady-state updates (Eq. \ref{eq:update_W_ss}) of Theorem \ref{theorem:grad_H}, is ensuring that saturated neurons refrain from updating their synaptic weights for the current data sample, as $\phi(\vv_i) - \phi(\vff_i)\approx 0$ in this case. 

\citet{meulemans2021credit} showed that including $\phi$ in the learning rule is a useful heuristic for improving the alignment of the DFC weight updates with the theoretically ideal update (which in our case would be the negative gradient of $\calH$), when the column space condition on $Q$ is not perfectly satisfied. Consequently, including $\phi$ also improves the performance in this case. When the column space condition on $Q$ is perfectly satisfied, including $\phi$ in the learning rule makes no difference.

\subsection{Steady-state weight updates vs. continuous weight updates} \label{app:steady_state_updates}
Our theoretical results assume that the weight update is performed at steady state, giving rise to the update defined in Eq. \eqref{eq:update_W_ss}. However, in practice, plasticity is always on, giving rise to the continuous plasticity dynamics defined in Eq. \eqref{eq:W_dynamics}. Following \citet{meulemans2021credit}, we show that when the steady state is quickly reached, the steady-state update \eqref{eq:update_W_ss} approximates the resulting continuous update \eqref{eq:W_dynamics} well. 

Assuming a separation of time scales between the network dynamics and the synaptic plasticity ($\tau_v, \tau_u \ll \tau_W$), the accumulated weight update resulting from the plasticity dynamics \eqref{eq:W_dynamics} is given by
\begin{align}
    \int_0^{t_1} \dd W_i = \int_0^{t_{\sss}} \dd W_i + \frac{t_1 - t_{\sss}}{\tau_W} \left(\phi(\vv_{i,\sss}) - \phi(\vff_{i,\sss}) \right) \vr_{i-1,\sss}^T.
\end{align}

Now, if the dynamics quickly settle towards their steady state, relative to the time interval the input is presented, i.e., $t_{\sss} \ll t_1$, the weight update is dominated by the second term in the RHS of the above equation, which corresponds to the steady-state weight update with $\phi$ included. Furthermore, the controller quickly pushes the activity close to its steady state, with possibly some oscillations around it, so the first term $\int_0^{t_{\sss}} \dd W_i$ will approximate the steady state well.

In conclusion, the Strong-DFC weight update resulting from the plasticity dynamics described in Eq. \eqref{eq:W_dynamics} will approximate the steady-state weight update $\left(\phi(\vv_{i,\sss}) - \phi(\vff_{i,\sss}) \right) \vr_{i-1,\sss}^T$. Combined with the arguments of Appendix \ref{app:nonlinearity_update}, we see that the Strong-DFC weight update approximates the steady-state update of Theorem \ref{theorem:grad_H} well.

\subsection{The minimizing control framework is compatible with various controller types} \label{app:other_controllers}
In Section \ref{sec:minimizing_control}, we introduced our theory on the minimizing control framework tailored towards a proportional intergral (PI) controller used by DFC. However, our theory can easily be extended to other controller types that reach the correct steady state. 

In the proof of Theorem \ref{theorem:grad_H}, the only assumption on the controller is its steady state equation \eqref{eq_app:controller_steady_state}, given by Eq. \eqref{eq_app:controller_steady_state}:
 \begin{align*}
    F(\vu_{\sss}, \vt) \triangleq \ve{e}_{\sss}(\vu_{\sss}, \vt) - \tilde{\alpha} \vus = 0.
 \end{align*}

Therefore, Theorem \ref{theorem:grad_H} applies to all stable controllers that give rise to this same steady-state equation, $F(\vu_{\sss}, \vt)$.

For example, pure proportional control, $\vu(t) = k\ve{e}(t)$ gives rise to the same steady-state equation $F(\vu_{\sss}, \vt)$, with $\talpha = \frac{1}{k}$. For pure integral control, we have the same steady-state equation $F(\vu_{\sss}, \vt)$ but with $\talpha = \alpha$. If we add derivative control to any of the above controller types, we reach the same steady-state equation as the derivative of the error vanishes at steady state. 

In conclusion, Theorem \ref{theorem:grad_H} is compatible with all possible combinations of the proportional-integral-derivative (PID) controller framework, except for the pure derivative control. Furthermore, all other controller types that satisfy the steady-state equation $F(\vu_{\sss}, \vt)$ are also compatible with Theorem \ref{theorem:grad_H}.

\subsection{Adapting the minimizing control framework for classification with softmax and cross-entropy loss} \label{app:softmax}
For classification with neural networks, one uses conventionally a one-hot label $\vrltrue$ in combination with a cross-entropy loss and a softmax layer as output of the network. In our minimizing control framework, the controller drives the network to perfectly match the output target $\vrltrue$. However, a softmax layer only results in a one-hot vector if it has an infinite input. To overcome this challenge, we use soft targets which have $a$ on its label entry and $\frac{1-a}{n_L-1}$ on the other entries, with $a$ close to 1 (e.g., 0.99).

A nice property of the combination of the cross-entropy loss with the softmax output is that the curvature of the exponentials in the softmax cancel out the curvature of the logarithms in the cross-entropy loss, leading to clean gradients that do not saturate easily. However, if our controller uses as control error $\ve{e}(t) = \vrl^* - \vrl(t)$, with $\vrl^*$ the soft target and $\vrl(t)$ the softmax output, it cannot make use of the cross-entropy loss to provide clean gradients. To overcome this challenge, we use a linear output layer, $\vrl$, and absorb the softmax into the cross-entropy loss, leading to the combined loss:
\begin{align}
    \calL^{\mathrm{combined}} = -\sum_{b=1}^B \ve{r}_L^{\mathrm{true},(b)T}\log\big(\mathrm{Softmax}(\ve{r}_L^{(b)})\big).
\end{align}

Next, we slightly adapt the minimizing control framework to include this combined loss in a principled way. Currently, the controller dynamics is given by
\begin{align}
    \vu(t) &= k \ve{e}(t) + \uint(t), \quad \tau_u \ddt \uint(t) = \ve{e}(t) - \alpha \uint(t)\\
    \ve{e}(t) &= \vrl^* - \vrl(t).
\end{align}
Note that $\ve{e}(t)$ can be interpreted as $-\pder{\vrl}{\mathcal{L}}\rvert_{\vrl=\vrl(t)}$, with $\mathcal{L} = \frac{1}{2}\snorm{\vrl^* - \vrl(t)}_{2}^{2}$. Now, if we use another loss $\mathcal{L}$ (i.e., $\calL^{\mathrm{combined}}$), we can generalize the control error $\ve{e}(t)$ to 
\begin{align}
    \ve{e}(t) = -\left.\pder{\vrl}{\mathcal{L}}\right\rvert_{\vrl=\vrl(t)}^T.
\end{align}
For the cross entropy loss with the softmax, this boils down to 
\begin{align}
    \ve{e}(t) = -\left.\pder{\vrl}{\mathcal{L}}\right\rvert_{\vrl=\vrl(t)}^T = \ve{p}^* - \mathrm{Softmax}(\vrl(t)),
\end{align}
with $\ve{p}^*$ the soft target.

\section{Stability analysis} \label{app:stability}
In this section, we build further upon the stability analysis results of \citet{meulemans2021credit}, which are applicable to Strong-DFC with minimal adaptations. 

As a first step, we investigate a pure integral controller ($k=0$) and assume a separation of timescales $\tauv \ll \tau_u$, i.e., we can replace the network dynamics by a deterministic function and only retain the controller dynamics. This provides us with the following controller dynamics:
\begin{align}
    \tau_u \dot{\vu}(t) = \ve{e}\left(\vu(t)\right) - \alpha \vu(t),
\end{align}
where the error, $\ve{e}$, can be seen as a deterministic function of the control input, $\vu(t)$ (for simplicity, we omit the dependence on the network parameters and input, as they are considered fixed during the controller dynamics). To investigate the local stability at steady state, we consider the eigenvalues of the Jacobian matrix of these dynamics evaluated at steady state. Using Eq. \eqref{eq_app:jacobian_steady_state_controller}, we have that:
\begin{align}
    \frac{\partial}{\partial \vu} \left.\left(\ve{e}\left(\vu\right) - \alpha \vu\right)\right\vert_{\vu=\vus} = -\left(\Js Q + \alpha I\right).
\end{align}
Hence, the condition for local stability around the steady state is that all eigenvalues of $-\left(\Js Q + \alpha I\right)$ should be negative, resulting in Condition \ref{con:stability}.

When considering the proportional integral controller without a separation of timescales, the local stability condition becomes harder to interpret. Following a similar derivation as proposed by \citet{meulemans2021credit}, we reach the following Jacobian of the network-controller dynamics: 
\begin{align}
    A_{\mathrm{PI}} = - 
    \begin{bmatrix}
            \frac{1}{\tau_v}(I-W\phi') & -\frac{1}{\tau_v} Q \\
            \frac{1}{\tilde{\tau}_u}S\phi' - \frac{k}{\tau_v}S\phi' (I-W\phi')& \frac{k}{\tau_v}S\phi'Q  \frac{\tilde{\alpha}}{\tilde{\tau}_u}I,
        \end{bmatrix}
\end{align}
with $\tilde{\tau}_u \triangleq \frac{\tau_u}{1+\alpha k}$, $\talpha \triangleq \frac{\alpha}{1+ \alpha k}$, $\phi' \triangleq \left.\pder{\vv}{\phi(\vv)}\right\vert_{\vv=\vv_{\sss}}$, and
\begin{align}
    W &\triangleq \begin{bmatrix}
    0 & 0 & 0&  \hdots  & 0 \\
    W_2 & 0& 0 &\hdots & 0 \\
    0 & W_3 & 0 &\hdots & 0 \\
    0 & 0 & \ddots & \hdots & 0 \\
    0 & 0 & \hdots & W_L & 0
    \end{bmatrix}\\
    S & \triangleq \begin{bmatrix} 0& \hdots & 0 & I \end{bmatrix}
\end{align}
The condition for local stability is now that all eigenvalues of $A_{\mathrm{PI}}$ are negative. However, this condition provides no straightforward interpretations for defining conditions on the feedback weights, $Q$, such that local stability is achieved. One can apply Gershgoring's circle theorem for finding sufficient conditions on $\Js$ and Q to ensure local stability, however, the resulting conditions are too conservative and hence not usable for designing feedback weight learning rules.

\section{Proofs and discussion for Section \ref{sec:fb_learning}}\label{app:fb_learning}
\subsection{Stochastic dynamics for the single-phase learning setting}\label{app:stochastic_dynamics}
In this section, we provide a detailed description of the stochastic dynamics used for training the forward and feedback weights in a single phase. The network and controller obey the following dynamics:
\begin{align}\label{eq_app:noisy_dynamics_v}
    \tau_v \ddt \vv_i(t) &= -\vv_i(t) + W_i \phi\big(\vv_{i-1}(t)\big) + Q_i \vu(t) + \sigma \veps_i(t)\\
    \tau_u \ddt \uint(t) &= \vrl^* - \vrl(t) -\alpha \uint(t) \label{eq_app:noisy_dynamics_uint}\\
    \vu(t) &= k \big(\vrl^* - \vrl(t)\big) + \uint(t) \label{eq_app:noisy_dynamics_u}\\
\end{align}
with $\sigma^2$ the noise variance and $\veps(t)$ exponentially filtered white noise, with time constant $\taue$:
\begin{align}
    \veps_i(t) = \frac{1}{\taue}\int_{-\infty}^{t}\exp\big(-\frac{1}{\taue}(t-t')\big)\vex_i(t')\dd t',
\end{align}
where $\vex$ is white noise. The time constant $\taue$ determines the auto-correlation of $\veps$. For simplicity, we take that the noise $\veps$ enters solely in the feedback compartment:
\begin{align}
    \vfb_i(t) = Q_i\vu(t) + \sigma \veps_i(t).
\end{align}
However, all the arguments and theorems of the following sections also apply for noise entering all compartments.

Now, we define the following error-based learning rule for the forward weights and anti-Hebbian learning rule for the feedback weights: 
\begin{align}
    \tau_W \ddt W_i(t) &= \Big(\phi\big(\vv_i(t)\big) - \phi\big(W_i\vr_{i-1}(t)\big)\Big) \bar{\vr}_{i-1}^T\\
    \tau_Q \ddt Q_i(t) &= -\left(1+\frac{\tau_v}{\taue}\right)^{L-i}\vfb_i(t)\vut(t)^T - \beta Q_i \label{eq:Q_dynamics}\\
    \vut(t) &= \vu(t) - \vub(t)\\
    \vub(t) &= \frac{1}{\tau_f}\int_{0}^t \exp\big(\frac{1}{\tau_f}(t-t')\big)\vu(t')\dd t'\\
    \bar{\vr}_i(t) &= \frac{1}{\tau_f}\int_{0}^t \exp\big(\frac{1}{\tau_f}(t-t')\big)\vr_i(t')\dd t'
\end{align}
with $\tau_f$ the filtering time constant used to high-pass filter $\vu$ and low-pass filter $\vr_i$. 

\subsection{Intuition behind the feedback weight learning rule}
Here we expand on the intuition behind our feedback weight learning rule. The central technique, inspired by the weight mirroring method \citep{akrout2019deep} and DFC \citep{meulemans2021credit}, is to inject white noise in the network that carries information about the network Jacobian, $\Js$, towards the output. If we now measure the correlation of the output noise w.r.t. the injected noise in the layers, we can recover the information on $\Js$. Below, we summarize the intuitive explanation on a normal feedforward neural network, provided by \citet{meulemans2021credit}. We end with explaining how to go from this intuitive explanation to our single-phase feedback weight plasticity, and how it differs from the original feedback weight plasticity of DFC. 

Now, let us consider a normal feedforward neural network $\vr_i = \phi(\vv_i) =  \phi(W_i \vr_{i-1})$. We perturb each layer with white noise $\sigma \vex$, which is propagated forward through the network:
\begin{align}
    \vvt_i &= W_i \phi(\vvt_{i-1}) + \sigma \vex_i \\
    \tilde{\vr}_0 &= \vr_0  
\end{align}
Assuming $\sigma$ is small, we can approximate the network output as:
\begin{align}
    \tilde{\vr}_L \approx \vrl + \sigma J \vex,
\end{align}
with $J=\begin{bmatrix}\pder{\vv_1}{\vrl}, & \hdots & ,\pder{\vv_L}{\vrl}   \end{bmatrix}$ and $\vex$ the concatenation of all $\vex_i$. Now, we can use an output error $\ve{e} = \vrl - \tilde{\vr}_L$ to extract the noise, and use this in a feedback learning rule $\Delta Q = -\vex \ve{e}^T - \beta Q$. In expectation, this update rule results in 
\begin{align}
    \bbE [\Delta Q] = \sigma^2 J^T - \beta Q.
\end{align}
Hence, we see that this simple update rule drives the feedback weights to align with $J^T$.

Although the above explanation grasps the intuition behind the feedback learning rule \eqref{eq:Q_dynamics_scaled}, there are some important differences when we consider the continuous single-phase plasticity dynamics.
\begin{enumerate}
    \item The pre-synaptic plasticity signal of the feedback weight update \eqref{eq:Q_dynamics_scaled} is the high-pass filtered control signal $\vuhp$ instead of the output noise we used before. When training the feedback and forward weights simultaneously, $\vu$ consists of two parts: (i) the control signal used to drive the network in expectation towards the output target $\vrl^*$, and (ii), the output noise fluctuations integrated over time by the controller. As we are interested in the noise fluctuations, we need to extract them, which we do by using the high-pass filtered version of $\vu$. 
    \item The postsynaptic plasticity signal is the activity in the feedback compartment $\vfb_i = Q_i \vu + \veps_i$, instead of purely the noise. As $Q_i \vu$ will correlate with $\vuhp$, and $\vuhp$ contains the integrated noise fluctuations over time, instead of the direct noise fluctuations, the feedback weights $Q$ will not align exactly with $J^T$, but with $J^TM$ instead, where $M$ is a positive definite matrix.
    \item To incorporate noise in the continuous dynamics of Strong-DFC, we need to use stochastic differential equations (SDEs), instead of discrete noise perturbations. Unlike discrete feedforward neural networks, the network dynamics do not propagate the layer activity instantaneously towards the output, but have a delay proportional to $\tau_v$. As white noise has no correlation in time, naively correlating the output noise fluctuations with the current layer noise would result in a signal of zero due to the time delays. In order to have correlations over time, we use exponentially filtered white noise $\veps$ with time constant $\taue$, instead of pure white noise $\vex$. As the time delay between the output noise fluctuations and the layer noise increases for upstream layers, we need to compensate for the decreasing correlation signal by scaling the plasticity updates with $\left(1 + \frac{\tauv}{\taue}\right)^{L-i}$, as is done in Eq. \eqref{eq:Q_dynamics_scaled}.
\end{enumerate}
Although our single-phase plasticity dynamics \eqref{eq:Q_dynamics_scaled} are closely related to the two-phase plasticity dynamics of DFC introduced by \citet{meulemans2021credit}, and these authors discussed briefly the possibility of having single-phase plasticity dynamics, our plasticity dynamics introduce some important advances, both in theory and practice. 
\begin{enumerate}
    \item Our minimizing control framework explicitly motivates learning the forward and feedback weights simultaneously, as the feedback weights are trained to satisfy Condition \ref{con:col_space}, which is evaluated at the steady state of the network for training the forward weights. By contrast, the theory for DFC in the weak-nudging regime requires the feedback weights to be trained at the network steady state without influence from the controller, and therefore different from the network steady state used for training the forward weights \citep{meulemans2021credit}.
    \item In their theoretical results on the feedback learning rule, \citet{meulemans2021credit} require that the controller leakage $\alpha$ is big. However, this cannot be combined with learning the feedback and feedforward weights simultaneously, as for training the feedforward weights, we need $\alpha \rightarrow 0$. In our new theoretical results, we overcome this important challenge. 
    \item In their theoretical results on the feedback learning rule, \citet{meulemans2021credit} assume instant network dynamics ($\tau_v \rightarrow 0$). However, in practice, a finite network time constant is used, giving rise to delays in the propagation of noise fluctuations to the output, leading to weaker correlation signals for upstream layers. Our new theory embraces finite time constants and investigates in detail the required relation between the various time constants $\tauv$, $\taue$, $\tau_u$, and $\tau_f$. As a result, we introduce the scaling term $\left(1 + \frac{\tauv}{\taue}\right)^{L-i}$ which improves the feedback learning significantly.
\end{enumerate}

\subsection{Proof of Theorem \ref{theorem:fb_learning}} \label{app:proof_fb_learning}
In the following, we prove Theorem \ref{theorem:fb_learning}, which we state in its full form below.
\begin{theorem}
Assume a separation of timescales $\tauv, \taue \ll \ \tau_u \ll \tau_f \ll \tau_Q$, integral control ($k=0$), stable network dynamics and small noise perturbations $\sigma \ll \snorm{\vv_i}$. Then, for a fixed data sample, the feedback plasticity dynamics \eqref{eq:Q_dynamics_scaled} let the first moment of $Q$ converge approximately towards the following configuration that satisfies Conditions \ref{con:col_space} and \ref{con:stability}:
\begin{align}\label{eq_app:fb_alignment}
    \mathbb{E}[Q_{\sss}] \approx \Js^T M,
\end{align}
with $M$ a positive definite symmetric matrix. 
\end{theorem}

\begin{proof}
We consider the noisy dynamics defined in Eq. (\ref{eq_app:noisy_dynamics_v}-\ref{eq_app:noisy_dynamics_u}), with $k=0$. 
Let us define $\vvbs$ and $\vubs$ as the steady state of the mean trajectory of the noisy dynamics, and $\vvt \triangleq \vv - \vvbs$ and $\vut \triangleq \vu - \vubs$ as the (noisy) perturbations around the mean trajectories. 
As $\sigma \ll \snorm{\vv_i}$ and the dynamics are stable (hence, contracting), we have that $\snorm{\vvt} \ll \snorm{\vv}$ and $\snorm{\vut} \ll \snorm{\vu}$, allowing us to do a first-order Taylor expansion of the dynamics, which for simplicity we assume to be exact:
\begin{align}
    \tau_v \ddt \vvt(t) &= -\vvbs + W \phi\big(\bar{\vv}_{\sss}\big) + U \vr_0 + Q \vubs(t) - \vvt(t) + W \phi' \vvt(t) + Q \vut(t) + \sigma \veps(t)\\
    &= - \vvt(t) + W \phi' \vvt(t) + Q \vut(t) + \sigma \veps(t)\\
    \tau_u \ddt \vut(t) &= \vrl^* - S\phi(\vvbs) -\alpha  \vubs - S \phi' \vvt(t) - \alpha \vut(t) \\
    &= - S \phi' \vvt(t) - \alpha \vut(t)
\end{align}
where we define $\phi' \triangleq \left.\pder{\vv}{\phi(\vv)}\right\vert_{\vv=\vvbs}$,
\begin{align}
    W &\triangleq \begin{bmatrix}
    0 & 0 & 0&  \hdots  & 0 \\
    W_2 & 0& 0 &\hdots & 0 \\
    0 & W_3 & 0 &\hdots & 0 \\
    0 & 0 & \ddots & \hdots & 0 \\
    0 & 0 & \hdots & W_L & 0
    \end{bmatrix}\\
    U &\triangleq \begin{bmatrix} W_1^T & 0 & \hdots & 0 \end{bmatrix}^T\\
    S & \triangleq \begin{bmatrix} 0& \hdots & 0 & I \end{bmatrix}
\end{align}
and use that $-\vvbs + W \phi\big(\bar{\vv}_{\sss}\big) + U \vr_0 + Q \vubs(t) = 0$ and $\vrl^* - S\phi(\vvbs) -\alpha  \vubs=0$ at the steady state of the mean trajectory. Similar to $\Delta \vv_i$, we define $\dvvt_i \triangleq \vvt_i - W_i \phi'_{i-1} \vvt_{i-1}$ with $\phi'_i \triangleq \left.\pder{\vv_i}{\phi(\vv_i)}\right\vert_{\vv=\vvbsi}$. As $\dvvt_1 = \vvt_1$ and $\vvt_i = \dvvt_i + W_i \phi'_{i-1} $, we have that 
\begin{align}
    S \phi' \vvt = \phi'_L \vvt_L = \Js \dvvt,
\end{align}
with $J_{\mathrm{ss}} \triangleq [\frac{\partial \vr_L}{\partial \vv_1}, ...,\frac{\partial \vr_L}{\partial \vv_L}]$ evaluated at $\vvbs$. Using $\ddt \dvvt = (I-W\phi')\ddt \vvt$ and $\vgamma(t) \triangleq [\dvvt(t)^T, \vut(t)^T]^T$ to simplify notation we get
\begin{align}
    \ddt \vgamma(t) = -\underbrace{\begin{bmatrix}
    \frac{1}{\tauv}(I - W\phi') & -\frac{1}{\tauv}(I-W\phi')Q \\
    \frac{1}{\tau_u} \Js & \frac{\alpha}{\tau_u} I 
    \end{bmatrix}}_{\triangleq A}
    \vgamma(t) + \underbrace{\begin{bmatrix} 
    \frac{\sigma}{\tau_v} (I-W\phi') \\ 0
    \end{bmatrix}}_{\triangleq B}
    \veps(t)
\end{align}

We solve this set of linear time-invariant differential equations using the variation of constants method \citep{sarkka2019applied}, taking $\vvt(t_0) = 0$ and $\vut(t_0) = 0$ (i.e., we consider that the network and controller dynamics are already converged at $t_0$). 
\begin{align}\label{eq_app:solution_gamma}
    \vgamma(t) = \int_{t_0}^t \exp\big(-A(t-t')\big)B\veps(t') \dd t'.
\end{align}

Now that we solved the differential equations for $\vvt$ and $\vut$, we turn our attention to the feedback weight plasticity rule \eqref{eq:Q_dynamics_scaled}. We start by investigating $\mathbb{E}[\vfb(t) \vut(t)^T]$. As the noise $\veps$ enters through the apical compartment, we have that:
\begin{align}
    \vfb(t) &= Q\vu(t) + \sigma\veps(t) = Q\vubs + Q\vut(t) + \sigma\veps(t) \\
    \mathbb{E}[\vfb(t) \vut(t)^T] &= Q\vubs \mathbb{E}[\vut(t)^T] + Q \mathbb{E}[\vut(t) \vut(t)^T] + \sigma \mathbb{E}[\veps(t) \vut(t)^T] \label{eq_app:expectation_vfb_vut}
\end{align}
The first term of Eq. \eqref{eq_app:expectation_vfb_vut} is equal to zero, as $\veps$ and consequently $\vut$ are zero-mean. We now turn our attention to the third term. Using $S_u \triangleq \begin{bmatrix} 0 & I \end{bmatrix}$ and $\vut = S_u \vgamma$, we have that:
\begin{align}
    \mathbb{E}[\vut(t) \veps(t)^T] &= S_u \mathbb{E}[\vgamma(t) \veps(t)^T] = S_u \int_{t_0}^t \exp\big({-}A(t-t')\big)B \mathbb{E}[\veps(t')\veps(t)^T]\dd t' \\
    &= \frac{1}{2\taue}S_u \int_{t_0}^t \exp\big({-}(A+ \frac{1}{\taue}I)(t-t')\big)B\dd t',
\end{align}
for which we used $\mathbb{E}[\veps(t')\veps(t)^T] = \frac{1}{2\taue} \exp\big(\frac{1}{\taue}|t-t'|\big)$. Using $\int_0^T\exp(At)\dd t = \big(\exp(AT) - I\big) A^{-1}$ and assuming stability and $t-t_0 \gg \tau_u$, which is justified by the condition $\tau_Q \gg \tau_u$, we can solve the previous integral:
\begin{align}
    \mathbb{E}[\vut(t) \veps(t)^T] = \frac{1}{2\taue}S_u \big(A + \frac{1}{\taue} I\big)^{-1} B.
\end{align}
Using the solution for the inverse of $2\times 2$ block matrices of \citet{lu2002inverses}, we get
\begin{align} \label{eq_app:scaled_jac}
    &\mathbb{E}[\vut(t) \veps(t)^T] =\\
    &{-}\frac{1}{2\taue} \left(\frac{1}{\taue} + \frac{\alpha}{\tau_u}\right)^{-1} \frac{1}{\tau_u} \Js \left(\frac{1}{\taue}I + \frac{1}{\tauv}\left(I-W\phi'\right) + \left(\frac{1}{\taue} + \frac{\alpha}{\tau_u}\right)^{-1}\frac{1}{\tauv}\left(I-W\phi'\right) Q\frac{1}{\tau_u} \Js\right)^{-1} \frac{\sigma}{\tauv} \left(I -W\phi'\right)
\end{align}
Now, using the condition $\tauv, \taue \ll \tau_u$, we can approximate the above with:
\begin{align}
    \mathbb{E}[\vut(t) \veps(t)^T] &\approx {-} \frac{\sigma}{2\tau_{u}} \Js \left(\left(1 + \frac{\tauv}{\taue}\right)I - W\phi' \right)^{-1}\left(I-W\phi'\right)
\end{align}
Using Lemma \ref{lemma_app:J_a}, we have that:
\begin{align}
    \mathbb{E}[\vut(t) \veps(t)^T] &\approx {-} \frac{\sigma }{2\tau_{u}} \left. \begin{bmatrix} 
    \left(1 + \frac{\tauv}{\taue}\right)^{-L}\pder{\vv_1}{\vrl} & \left(1 + \frac{\tauv}{\taue}\right)^{-(L-1)}\pder{\vv_2}{\vrl} & \hdots & \left(1 + \frac{\tauv}{\taue}\right)^{-1}\pder{\vv_L}{\vrl}
    \end{bmatrix} \right\vert_{\vv=\vv_{\sss}}
\end{align}

Now we turn our attention towards the second term of Eq. \eqref{eq_app:expectation_vfb_vut}. As $\bbE[\vut(t) \vut(t)^T]$ is a covariance matrix, it is positive semi-definite. 

Combining Eq. \eqref{eq:Q_dynamics_scaled} with Eq. \eqref{eq_app:scaled_jac} and the above insights, we get
\begin{align}
    \tau_Q \ddt \bbE[Q(t)] \triangleq \tau_Q \ddt \bar{Q}(t) = \frac{\sigma^2 }{2\tau_{u}}\Js^T - \bar{Q} (\tilde{M} + \beta I),
\end{align}
with $\tilde{M}$ a positive semi-definite matrix. We used the fact that the layer-wise scaling with $\left(1+\frac{\tauv}{\taue}\right)^{L-i}$ in Eq. \eqref{eq:Q_dynamics_scaled} cancels out the scaling factors $\left(1+\frac{\tauv}{\taue}\right)^{-(L-i)}$ in Eq. \eqref{eq_app:scaled_jac}. Using $M\triangleq \tilde{M} + \beta I$, the stable point of the above dynamics is 
\begin{align}
    \bar{Q}_{\sss} = \frac{\sigma^2}{\tau_{u}}\Js^T M^{-1}.
\end{align}
As $M$ is positive definite, the dynamics are stable and the above steady state is reached. Furthermore, as $\Js^T M^-1$ is a permutation of the columns of $\Js^T$, it satisfies Condition \ref{con:col_space}; and as $M^{-1}$ is positive definite, it satisfies Condition \ref{con:stability}.

Now the only thing left to show is that $\vuhp \triangleq \vu - \vulp \overset{?}{=} \vut$, with $\vulp$ the low-pass filtered version of $\vu$:
\begin{align}
    \vulp(t) = \frac{1}{\tau_f} \int_{t_0}^{t} \exp\left({-}\frac{1}{\tau_f}(t-t')\right)\vu(t') \dd t'.
\end{align}
As we have that $\tau_f \gg \tau_u,\taue, \tauv$, the low-pass filtering will remove all fluctuations and we have that $\vulp = \vubs$. Consequently, we have that $\vuhp = \vu - \vubs = \vut$, thereby concluding the proof.
\end{proof}

\begin{lemma}\label{lemma_app:J_a}
For some scalar $a\neq0$, we have that 
\begin{align}
    J \left(aI -W\phi'\right)^{-1} \left(I - W\phi'\right) = \left. \begin{bmatrix} 
    a^{-L}\pder{\vv_1}{\vrl} & a^{-(L-1)}\pder{\vv_2}{\vrl} & \hdots & a^{-1}\pder{\vv_L}{\vrl}
    \end{bmatrix} \right\vert_{\vv=\vv'}
\end{align}
with $J$ and $\phi'$ evaluated at some activity pattern $\vv'$.
\end{lemma}
\begin{proof}
First, we note that $\left(aI -W\phi'\right)$ is of full rank for $a\neq 0$, as it is a triangular matrix with $a$ on its diagonal. The inverse of $\left(aI -W\phi'\right)$ is given by:
\begin{align}
    \left(aI -W\phi'\right)^{-1} = \begin{bmatrix}
    a^{-1} I & 0&0&\hdots&0 \\
    a^{-2} J_{2,1} & a^{-1} I & 0 & \hdots & 0 \\
    a^{-3} J_{2,1} & a^{-2} J_{3,2} & a^{-1} I & \hdots & 0 \\
    \vdots& \ddots & \ddots & \ddots & \vdots \\
    a^{-L} J_{L,1} & \hdots & \hdots & a^{-2} J_{L,L-1} & a^{-1} I
    \end{bmatrix}
\end{align}
with $J_{i,j} \triangleq \pder{\vv_j}{\vv_i}$. One can easily verify this by computing $\left(aI -W\phi'\right) \left(aI -W\phi'\right)^{-1} = I$ upon noting that $J_{i,i-1} = W_i \phi'_{i-1}$. Next, we have that:
\begin{align}
    \left(aI -W\phi'\right)^{-1} \left(I - W\phi'\right) = 
    \begin{bmatrix}
    a^{-1} I & 0&0&\hdots&0 \\
    (a^{-2}-a^{-1}) J_{2,1} & a^{-1} I & 0 & \hdots & 0 \\
    (a^{-3}-a^{-2}) J_{2,1} & (a^{-2}-a^{-1}) J_{3,2} & a^{-1} I & \hdots & 0 \\
    \vdots& \ddots & \ddots & \ddots & \vdots \\
    (a^{-L}-a^{-(L-1)}) J_{L,1} & \hdots & \hdots & (a^{-2}-a^{-1}) J_{L,L-1} & a^{-1} I
    \end{bmatrix}
\end{align}
Using $J = [\pder{\vv_1}{\vrl}, \pder{\vv_2}{\vrl}, \hdots, \pder{\vv_L}{\vrl}]$, we have 
\begin{align}
    J \left(aI -W\phi'\right)^{-1} \left(I - W\phi'\right) =  \begin{bmatrix} 
    a^{-L}\pder{\vv_1}{\vrl} & a^{-(L-1)}\pder{\vv_2}{\vrl} & \hdots & a^{-1}\pder{\vv_L}{\vrl},
    \end{bmatrix}
\end{align}
thereby concluding the proof.
\end{proof}

\subsection{Noise can bias the feedforward weight updates} \label{app:noise_bias}
As observed by \citet{meulemans2021credit}, when we inject noise into the dynamics while training the forward weights, the resulting noise correlations can bias the forward weight updates. This is a general issue for methods using error feedback for learning while having realistic noise dynamics, as discussed in more detail in App. C.4 in \citet{meulemans2021credit}. Here, we investigate how these issues manifest themselves in Strong-DFC and how we can solve them.

The forward weight update is given by Eq. \eqref{eq:W_dynamics}. If the noise fluctuations in the postsynaptic term $\phi(\vv_i(t)) - \phi(W_i\vr_{i-1}(t))$ are correlated to the noise fluctuations in the presynaptic term $\vr_{i-1}(t)$, the expected forward weight update contains this extra correlation term and is hence biased.

The difference $\phi(\vv_i(t)) - \phi(W_i\vr_{i-1}(t))$ is mainly determined by the feedback input $Q\vu(t)$. Using the notation of Section \ref{app:proof_fb_learning}, we divide the control signal of the converged dynamics in two parts: $\vu(t) = \vus + \vut(t)$, with $\vus$ the part of the control signal useful for learning the forward weights, and $\vut$ the noise fluctuations that are used for learning the feedback weights. These noise fluctuations are a function of the noise fluctuations of all network layers (e.g., see Eq. \ref{eq_app:solution_gamma}). Hence, the noise fluctuations of $\phi(\vv_i(t)) - \phi(W_i\vr_{i-1}(t))$ will contain a part of the noise fluctuations of $\vr_{i-1}(t)$, therefore causing the pre- and post-synaptic term in the plasticity rule to be correlated. 

A simple approach to remove the noise correlation in the plasticity rule for the forward weights, is either to remove the noise fluctuations in the pre- or post-synaptic term, or both. We choose to remove the noise fluctuations in the pre-synaptic term, by using a low-pass filtered version of $\vr_{i-1}$: 
\begin{align}
    \bar{\vr}_i(t) &= \frac{1}{\tau_f}\int_{0}^t \exp\big(\frac{1}{\tau_f}(t-t')\big)\vr_i(t')\dd t',
\end{align}
with $\tau_f$ the filtering time constant. This low-pass filtering can be interpreted as a leaky biophysical mechanism that accumulates the pre-synaptic input used for the plasticity of the synapse. As the pre-synaptic term of the forward plasticity dynamics no longer contains the noise fluctuations, the updates will no longer be biased.

\section{Simulation and algorithms} \label{app:simulation_algorithms}
\subsection{Simulating the dynamics}
We use the Euler-Maruyama method to simulate the stochastic differential equations (SDE), which simulates a (non)linear SDE 
\begin{align}
    \ddt \ve{x}(t) = \ve{f}(\ve{x}, t) + \ve{g}(\ve{x}, t) \vex(t)
\end{align}
as follows: 
\begin{align}
    \ve{x}[m+1] = \ve{x}[m] + \ve{f}(\ve{x}[m], m\Delta t) \Delta t + \ve{g}(\ve{x}[m], m\Delta t) \sqrt{\Delta t} \Delta \ves{\beta}_m,
\end{align}
with $\Delta \ves{\beta}$ a Gaussian random variable with distribution $\mathcal{N}(\ves{0}, I)$. Note that the noise term has stepsize $\sqrt{\Delta t}$ instead of $\Delta t$, as Brownian motion has a variance of $t$ and not $t^2$.

Applying the Euler-Maruyama method to the SDEs of Section \ref{app:stochastic_dynamics} defining the neural and controller dynamics gives us the following discrete update equations:
\begin{align}
\veps_i[m+1] &= \veps_i[m] + \frac{1}{\taue} \big(-\veps_i[m]\Delta t + \sqrt{\Delta t} \Delta \ves{\beta}_{i,m}\big)\\
\vfb_i[m] &= Q_i \vu[m] + \sigma\veps_i[m]\\
\uint[m+1] &= \uint[m] + (\ve{e}[m] - \alpha \uint[m])\frac{\Delta t}{\tau_u}\\
\vu[m] &= k \ve{e}[m] + \uint[m]\\
\vv_i[m+1] &= \vv_i[m] + (-\vv_i[m] + W_i \phi(\vv_{i-1}[m]) + \vfb_i[m])\frac{\Delta t}{\tauv}
\end{align}
And for the weight updates:
\begin{align}
\vub[m+1] &= \vub[m] + \frac{1}{\tau_f}(-\vub[m] + \vu[m])\Delta t\\
\bar{\vr}_i[m+1] &= \bar{\vr}_i[m] + \frac{1}{\tau_f}(-\bar{\vr}_i[m] + \vr_i[m])\Delta t\\
    \Delta W_i &= \frac{\Delta t}{\tau_W}\sum_m \big(\vr_i[m] - \phi(W_i\vr_{i-1}[m]) \big) \bar{\vr}_{i-1}[m]^T \\
    \Delta Q_i &= \frac{\Delta t}{\tau_Q}\left(1+\frac{\tau_v}{\taue}\right)^{L-i}\sum_m -\vfb_i[m](\vu[m]-\vub[m])^T - M_{\max}\frac{\Delta t}{\tau_Q}\beta Q_i
\end{align}
with $M_{\max}$ the total number of simulation steps.

\subsection{Pseudocode for the Strong-DFC algorithm}
Algorithm \ref{al:strong-DFC} provides the pseudocode for the Strong-DFC algorithm, making use of the discretized dynamics provided in the previous section. We use the Euler-Maruyama discretization of the SDEs with some slight modifications. 

First, in order to have direct control over the hyperparameter $\talpha$, we rewrite the controller dynamics as follows:
\begin{align} \label{eq_app:controller_dynamics_modified}
    \ve{u}(t) = \ve{u}^{\text{int}}(t) + k \ve{e}(t), \quad 
    \tau_u \ddt \ve{u}^{\text{int}}(t) = \ve{e}(t) - \tilde{\alpha} \ve{u}(t).
\end{align}
Note that the underlying dynamics for $\vu(t)$ are exactly the same as in Eq. \eqref{eq:controller_dynamics}. 

In order to better incorporate the layered structure of the network into the discrete network dynamics, we update a layer at timestep $m+1$ with the previous layer at timestep $m+1$ instead of $m$: $\vff_i[m+1] = W_i\phi(\ve{v}_{i-1}[m+1]) + \ve{b}_i$, with $\ve{b}_i$ the bias parameters of layer $i$. For small stepsizes $\Delta t$ this has almost no effect, but for bigger stepsizes it has the benefit of better reflecting the layered structure.

In order to better incorporate the controller loop into the discrete dynamics, we use the updated control signal $\vu[m+1]$ for updating the feedback compartment: $\vfb_i[m+1] = Q_i \ve{u}[m+1]$, such that the control error $\ve{e}[m]$ of the previous timestep is used for control instead of the error of two timesteps ago. For small stepsizes $\Delta t$ this has almost no effect, but for bigger stepsizes it has the benefit of better reflecting the control interaction.

Following the exponential filtering approach for the low-pass filtering of $\vr$ and $\vu$, we use the current value $[m+1]$ of $\vr$ and $\vu$ instead of the previous one prescribed by the forward Euler method. This reduces the time delay introduced by the low pass filtering.

\begin{algorithm}[tb]
   \caption{Pseudocode for the Strong-DFC algorithm on a single input sample}
   \label{al:strong-DFC}
\begin{algorithmic}
   \STATE \# Initialize layer activations to the feedforward activations and parameter update buffers to zero
    \FOR{$i$ in range($1,L$)}
\STATE $\vv_i[1] = \vv_i^-$ 
\STATE $\vr_i[1] = \vr_i^-$
\STATE $\Delta W_i = 0 $ 
\STATE $\Delta \ve{b}_i = 0$
\STATE $\Delta Q_i = 0 $ 
    \ENDFOR
    \STATE $\uint[1] = 0$ 
    \FOR{$m$ in range(1,$M_{\max}$)}
\STATE \# Update controller:
\IF{softmax classification with cross-entropy loss}
\STATE $\ve{e}[m] = \ve{p}^* - \mathrm{softmax}(\ve{r}_L[m])$ \COMMENT{with $\ve{p}^*$ the soft target and $\vrl$ a linear output layer}
\ELSIF{regression with L2 loss}
\STATE $\ve{e}[m] = \ve{r}_L^* - \ve{r}_L[m]$ 
\ENDIF
\STATE $\uint[m+1] = \uint[m] + \frac{\Delta t}{\tau_u} (\ve{e}[m] - \tilde{\alpha} \ve{u}[m])$ 
\STATE $\ve{u}[m+1] = \uint[m+1] + k \ve{e}[m] $ 
\STATE \# Low-pass filter $\vu$
\IF{$m==1$}
\STATE $\vub[m+1] = \vu[m+1]$ \COMMENT{initialize $\vub$}
\ELSE
\STATE $\vub[m+1] = \vub[m] + \frac{1}{\tau_f}(-\vub[m] + \vu[m+1])\Delta t$
\ENDIF
\STATE \# Update network:
\FOR{i in \text{range}(1,L)}
\STATE \# Sample noise to update the exponentially filtered noise variable
\STATE $\Delta \ves{\beta}_{i,m} \sim \mathcal{N}(0,I)$
\STATE $\veps_i[m+1] = \veps_i[m] + \frac{1}{\taue} \big(-\veps_i[m]\Delta t + \sqrt{\Delta t} \Delta \ves{\beta}_{i,m}\big)$
\STATE \# Inject noise in the feedback compartment and update the activations
\STATE $\vfb_i[m+1] = Q_i \vu[m+1] + \sigma\veps_i[m+1]$ 
\STATE $\vff_i[m+1] = W_i\phi(\ve{v}_{i-1}[m+1]) + \ve{b}_i$
\STATE $\vv_i[m+1] = \vv_i[m] + \frac{\Delta t}{\tau_v} ( -\vv_i[m] + \vff_i[m+1] + \vfb_i[m+1])$ \\
\STATE $\vr_i[m+1] = \phi(\vv_i[m+1])$ \\
\STATE \# low-pass filter $\vr_i$
\STATE $\bar{\vr}_i[m+1] = \bar{\vr}_i[m] + \frac{1}{\tau_f}(-\bar{\vr}_i[m] + \vr_i[m+1])\Delta t$

\STATE \# Buffer parameter updates:
\STATE $\Delta W_i = \Delta W_i + \big(\vr_i[m+1] - \phi(W_i\vr_{i-1}[m+1]) \big) \bar{\vr}_{i-1}[m+1]^T$ 
\STATE $\Delta \ve{b}_i = \Delta \ve{b}_i + \vr_i[m+1] - \phi(\vff_i[m+1])$ 
\STATE $\Delta Q_i = \Delta Q_i - \left(1+\frac{\tau_v}{\taue}\right)^{L-i}\vfb_i[m](\vu[m+1]-\vub[m+1])^T - \beta Q_i$ 
\ENDFOR
\ENDFOR
\STATE Update parameters with $\Delta W_i/M_{\max}$, $\Delta \ve{b}_i/M_{\max}$ and $\Delta Q_i/M_{\max}$ using an optimizer of choice

\end{algorithmic}
\end{algorithm}



\section{Experiments} \label{app:experiments}

\subsection{Alignment measures}

Below we describe how the measures of Fig. \ref{fig:toy:curves} were computed.

\paragraph{Alignment of Strong-DFC updates with $\mathcal{H}$} Fig. \ref{fig:toy:curves}B quantifies alignment between actual Strong-DFC updates and the gradient of $\mathcal{H}$. For this, we vectorize the update matrices and compute the angle between them in degrees.

\paragraph{Relative strength of feedback control} Fig. \ref{fig:toy:curves}C quantifies the magnitude of the feedback input compared to the forward drive, and is used to illustrate that our framework allows feedback to have comparable magnitudes to the forward input. It is calculated as follows:
\begin{align}
    \mathrm{ratio}_{\text{fb}/\text{ff}} = \frac{\|  Q\mathbf{u}  \|_F}{\| W\mathbf{r} \|_F},
\end{align}
where $F$ indicates the Frobenius norm.

\paragraph{Condition \ref{con:col_space}} Fig. \ref{fig:toy:curves}D quantifies the degree to which Condition \ref{con:col_space} is satisfied. It is computed through the Frobenius norm of the projection of $Q$ with the norm of $Q$:
\begin{align}
    \mathrm{ratio}_{\mathrm{Con1}} = \frac{\|P_{J_{ss}^T} Q\|_F}{\|Q\|_F}.
\end{align}

where the projection $P_{J_{ss}^T }Q$ is given by:
\begin{align}
    P_{J_{ss}^T} Q = J_{ss}^T (J_{ss}J_{ss}^T)^{-1} J_{ss} Q.
\end{align}
For an intuition of this metric, please refer to Appendix F.1 in \citet{meulemans2021credit}.

\subsection{Noise robustness measures}
\label{app:noise_robustness}

\begin{figure*}[h!]
\centering
\includegraphics[width=0.35\textwidth]{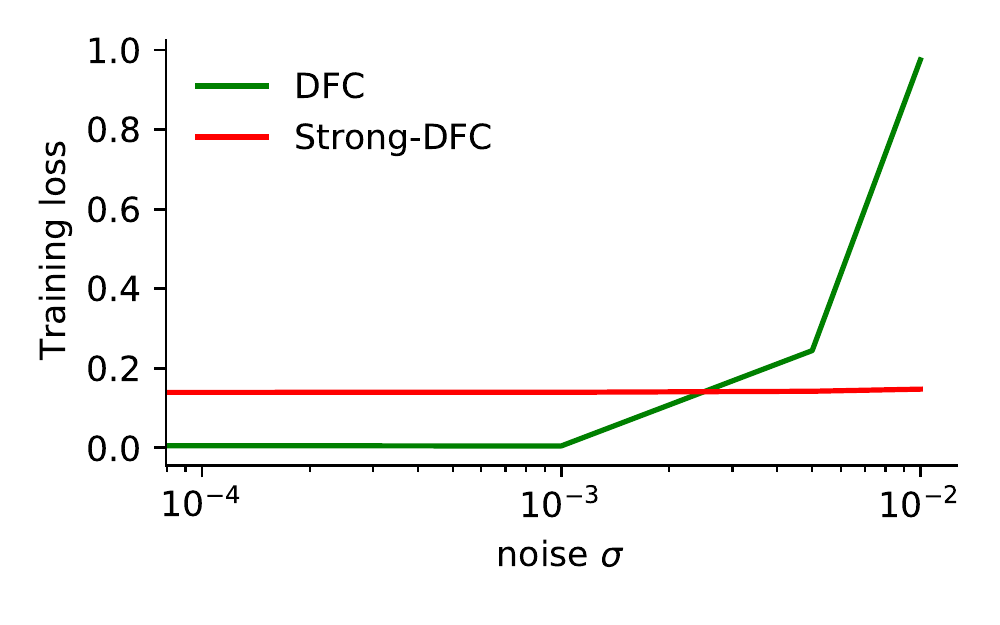}
\vspace{-0.3cm}
\caption{ \textbf{Noise robustness.} Training loss after 500 epochs (training not fully converged for Strong-DFC) for different noise magnitudes. The DFC results for high noise values are missing since its training was unstable.}
\label{fig:noise_robustness}
\end{figure*}

Below we further illustrate the robustness of Strong-DFC weight updates under noisy conditions. Recall that this is a direct consequence of the main novelty of Strong-DFC compared to standard DFC: the use of the true output target $\mathbf{r}_L^* = \mathbf{r}_L^{\mathrm{true}}$ throughout training. 
For this series of experiments (Figure \ref{fig:noise_robustness}) we vary the noise magnitude $\sigma$ added to the neural dynamics during training of the forward weights for DFC and Strong-DFC. For a fair comparison between both methods, we low-pass filter the presynaptic plasticity signal for debiasing the updates \eqref{eq:debiased-dW}. In a student-teacher regression setting with a nonlinear network of size 20-40-5, we found that the performance of Strong-DFC is more robust to a wide range of noise magnitudes compared to standard DFC. For non-noisy dynamics, we observed that standard DFC can descend the loss landscape faster compared to Strong-DFC in this setting, leading to a lower loss, as the Strong-DFC training has not yet fully converged after 500 epochs. For the results of Table~\ref{tab:my_label2} in the main manuscript, we used $\sigma=0.005$ for both methods.

\section{Additional Results}

\subsection{Training Strong-DFC without re-visiting samples}
\label{app:revisiting}

To investigate whether Strong-DFC requires consistency or re-visiting training samples multiple times to perform well, we performed the following experiment using the student-teacher setting. Using a teacher of size 30-10-10-10-5 and a student of size 30-50-50-50-5, we trained networks for 300 epochs, each containing 500 training samples. In the standard setting, the 500 training samples were identical across epochs. In the modified setting, each training sample was independently drawn from the training distribution, so that no sample was seen twice. This was achieved by feeding the teacher with a new set of randomly-generated input samples at each epoch. Therefore, training consisted of a total of 150000 samples, but in the modified setting all these samples were unique. The results obtained are presented in Table \ref{app:tab:retraining}.

\begin{table}[h]
    \centering
        \caption{Effect of sample reuse when training Strong-DFC.}
    \label{app:tab:retraining}
    \begin{tabular}{ c  c  c}
\toprule
Training setting  & Train loss & Test loss \\
\midrule
\midrule
Standard & $1.945^{\pm 0.001}$ & $2.224^{\pm 0.002}$  \\ 
Without sample reuse  &$1.964^{\pm 0.004}$ & $2.226^{\pm0.006}$  \\
\end{tabular}
\end{table}

The results obtained are extremely similar, indicating that Strong-DFC does not necessarily rely on re-visiting training samples many times to perform well. However, it is important to note that Strong-DFC needs i.i.d. samples and will not be immune to forgetting and interference issues when exposed to correlated sequences of training samples; solving these issues will likely require continual learning techniques, which is out of the scope for this work.